\documentclass[preprint,12pt]{elsarticle}

\makeatletter
\def\ps@pprintTitle{%
   \let\@oddhead\@empty
   \let\@evenhead\@empty
   \let\@oddfoot\@empty
   \let\@evenfoot\@empty
}
\makeatother

\usepackage{amssymb}
\usepackage{amsmath}

\usepackage{booktabs}
\usepackage{tabularx}
\usepackage{tabularray}
\usepackage{makecell}
\usepackage{hyperref}
\usepackage{subcaption}
\usepackage{float}
\usepackage{multirow} 
\usepackage{array}
\usepackage{adjustbox}
\usepackage{bm}
\usepackage{textcomp}
\usepackage{pifont}
\usepackage{comment}
\usepackage{overpic}

\begin{document}

\begin{frontmatter}



\title{DISARM++: Beyond scanner-free harmonization}

\author[aff:polimi]{Luca Caldera}
\author[aff:polimi]{Lara Cavinato}
\author[aff:genova-hospital]{Alessio Cirone}
\author[aff:unige-math,aff:unige-neuro]{Isabella Cama}
\author[aff:unige-math,aff:genova-hospital]{Sara Garbarino}
\author[aff:unibo-neuro,aff:bologna-irccs]{Raffaele Lodi}
\author[aff:besta-neuro]{Fabrizio Tagliavini}
\author[aff:besta-radio]{Anna Nigri}
\author[aff:brescia-fate]{Silvia De Francesco}
\author[aff:cattolica]{Andrea Cappozzo}
\author[aff:unige-math,aff:genova-hospital]{Michele Piana}
\author[aff:polimi,aff:ht-milan]{Francesca Ieva}
\author[aff:rin]{RIN-Neuroimaging Network}
\author[]{Alzheimer’s Disease Neuroimaging Initiative\footnote{Data used in preparation of this article were obtained from the Alzheimer’s Disease Neuroimaging Initiative (ADNI) database (adni.loni.usc.edu). As such, the investigators within the ADNI contributed to the design and implementation of ADNI and/or provided data but did not participate in analysis or writing of this report.}}

\affiliation[aff:polimi]{organization={MOX, Department of Mathematics, Politecnico di Milano},
            addressline={Via Bonardi 9},
            city={Milan},
            postcode={20133},
            state={Italy},
            country={Italy}}

\affiliation[aff:unige-math]{organization={Department of Mathematics, Università di Genova},
            addressline={Via Dodecaneso 35},
            city={Genova},
            postcode={16146},
            state={Italy},
            country={Italy}}

\affiliation[aff:unige-neuro]{organization={Department of Neuroscience, Rehabilitation, Ophthalmology, Genetics and Maternal-Child Sciences, Università di Genova},
            addressline={Largo Paolo Daneo 3},
            city={Genova},
            postcode={16132},
            state={Italy},
            country={Italy}}

\affiliation[aff:genova-hospital]{organization={IRCCS Ospedale Policlinico San Martino},
            addressline={Largo R. Benzi 10},
            city={Genova},
            postcode={16132},
            state={Italy},
            country={Italy}}

\affiliation[aff:unibo-neuro]{organization={Department of Biomedical and Neuromotor Sciences, University of Bologna},
            addressline={Via Zamboni 33},
            city={Bologna},
            postcode={40126},
            state={Italy},
            country={Italy}}

\affiliation[aff:bologna-irccs]{organization={IRCCS Institute of Neurological Science of Bologna, Bellaria Hospital},
            addressline={Via Altura 3},
            city={Bologna},
            postcode={40139},
            state={Italy},
            country={Italy}}

\affiliation[aff:besta-neuro]{organization={Unit of Neurology (V) and Neuropathology, IRCCS Istituto Neurologico Carlo Besta},
            addressline={Via Celoria 11},
            city={Milan},
            postcode={20133},
            state={Italy},
            country={Italy}}

\affiliation[aff:besta-radio]{organization={Neuroradiology Unit, IRCCS Istituto Neurologico Carlo Besta},
            addressline={Via Celoria 11},
            city={Milan},
            postcode={20133},
            state={Italy},
            country={Italy}}

\affiliation[aff:brescia-fate]{organization={Laboratory of Neuroinformatics, IRCCS Istituto Centro San Giovanni di Dio Fatebenefratelli},
            addressline={Via Pilastroni 4},
            city={Brescia},
            postcode={25125},
            state={Italy},
            country={Italy}}

\affiliation[aff:cattolica]{organization={Department of Statistical Sciences, Università Cattolica del Sacro Cuore},
            addressline={Largo Gemelli 1},
            city={Milan},
            postcode={20123},
            state={Italy},
            country={Italy}}

\affiliation[aff:ht-milan]{organization={Health Data Science Centre, Human Technopole},
            addressline={V.le Rita Levi-Montalcini 1},
            city={Milan},
            postcode={20157},
            state={Italy},
            country={Italy}}

\affiliation[aff:rin]{organization={RIN-Neuroimaging Network},
            addressline={Via Clericetti 2},
            city={Milan},
            postcode={20133},
            state={Italy},
            country={Italy}}
            

\begin{abstract}
Harmonization of T1-weighted MR images across different scanners is crucial for ensuring consistency in neuroimaging studies. This study introduces a novel approach to direct image harmonization, moving beyond feature standardization to ensure that extracted features remain inherently reliable for downstream analysis. Our method enables image transfer in two ways: (1) mapping images to a scanner-free space for uniform appearance across all scanners, and (2) transforming images into the domain of a specific scanner used in model training, embedding its unique characteristics.
Our approach presents strong generalization capability, even for unseen scanners not included in the training phase. We validated our method using MR images from diverse cohorts, including healthy controls, traveling subjects, and individuals with Alzheimer’s disease (AD). The model’s effectiveness is tested in multiple applications, such as brain age prediction ($R^2 \simeq 0.60 \pm 0.05$), biomarker extraction, AD classification (Test Accuracy $\simeq 0.86 \pm 0.03$), and diagnosis prediction (AUC $\simeq 0.95$). In all cases, our harmonization technique outperforms state-of-the-art methods, showing improvements in both reliability and predictive accuracy.
Moreover, our approach eliminates the need for extensive preprocessing steps, such as skull-stripping, which can introduce errors by misclassifying brain and non-brain structures. This makes our method particularly suitable for applications that require full-head analysis, including research on head trauma and cranial deformities. Additionally, our harmonization model does not require retraining for new datasets, allowing smooth integration into various neuroimaging workflows. By ensuring scanner-invariant image quality, our approach provides a robust and efficient solution for improving neuroimaging studies across diverse settings. The code is available at this \href{https://github.com/luca2245/DISARMpp_Harmonization.git}{link}. 
\end{abstract}


\end{frontmatter}



\section{Introduction}
\label{sec:introduction}
As brain Magnetic Resonance Imaging (MRI) datasets from various research centers become increasingly accessible, there is a growing opportunity to gain valuable insights into brain-related diseases. These insights have the potential to enhance medical practices by providing robust statistical evidence to be translated into clinical practice. However, variations in MRI data across different centers and scanners, due to unstandardized protocols, scanner- and acquisition-specific variabilities, can lead to significant inconsistencies in the extracted biomarkers, thereby affecting their repeatability and reproducibility. Differences in hardware, software configurations, calibration procedures, maintenance practices, and operators experience can cause MRI scanners to produce images with varying contrast, brightness, and spatial resolution, i.e., voxel intensity distribution. This variability, particularly in multicenter studies, can introduce confounding effects that compromise the reliability of the results \cite{takao2011effect,shinohara2017volumetric,zuo2019harnessing}. To address this issue, it is essential to harmonize MRI data across centers and scanners to ensure consistency and comparability of the datasets, thus strengthening the integrity of subsequent analyzes. By minimizing inter- and intra-scanner variability, harmonization enables researchers to reliably combine data from multiple sites, facilitating the development of robust machine learning and deep learning models that depend on large, high-quality datasets.

\section{Related Works}
\label{sec:related_works}
Several methodologies have been proposed to harmonize images and specifically multicenter MRI data. The approaches can generally be divided into two primary categories: feature-based and image-based approaches. 

\subsection{Feature-Based Approaches}
\label{subsec:feature_based}
Feature-based harmonization aims to align extracted features, such as cortical thickness, functional connectivity, or diffusion metrics, across batches rather than directly modifying the original images. A key method in this category is the widely used ComBat and its extensions, which leverage empirical Bayes frameworks to adjust for mean and variance shifts caused by batch effects \cite{torbati2021multi,radua2020increased}. These methods have been extensively validated across various datasets and data types. Typically, they rely on a linear model framework to harmonize features while preserving biologically meaningful variance. Extensions like ComBat-GAM \cite{pomponio2020harmonization} and CovBat \cite{chen2022mitigating} further refine these approaches by addressing nonlinear covariate effects and multivariate dependencies, respectively. Deep learning approaches have also been explored for feature-based harmonization. For example, Conditional Variational Autoencoders (CVAEs) have been applied to learn batch-invariant latent representations of imaging features, facilitating their reconstruction in a harmonized space \cite{an2022goal, moyer2020scanner, cavinato2023dual}. 

\subsection{Image-Based Approaches}
\label{subsec:image_based}
Image-based methods operate directly on raw imaging data, using machine learning techniques to adjust for batch effects at the voxel or pixel level, creating visual consistency across datasets collected from multiple centers. In the domain of image-based harmonization, several methods have been explored. These techniques can be categorized into three main frameworks: Transformers, Image-to-Image (I2I) translation, and Style Transfer. Transformers \cite{guo2022transformer} are a class of models built around a self-attention mechanism, which enables them to capture long-range dependencies within the data. However, despite their success in many applications, transformers have been noted for their limitations in retaining fine-grained, high-frequency details, which could be critical for accurately identifying biology- and pathology-related features. For instance, studies have shown that transformers may overlook important high-frequency information, thus potentially missing crucial signals relevant for medical imaging analysis \cite{wang2022anti}. Image-to-Image (I2I) translation involves mapping an input image to an output image in a way that preserves specific semantic properties while adjusting the image to match the target domain. The core of this method lies in generative models that are trained to produce images that resemble those drawn from the target distribution. I2I methods can be further classified according to the type of supervision (e.g., supervised, unsupervised) or the nature of the translation process (e.g., one-to-one, one-to-many, many-to-many) \cite{alotaibi2020deep}. Style Transfer \cite{liu2023style}, on the other hand, treats the harmonization process as a domain adaptation problem, focusing on transferring the style of an image from one dataset to another. In this fully unsupervised framework, the primary goal is to maintain the content of the original images while adjusting their style to align with the desired target. Image-based harmonization methods have already demonstrated considerable potential. Notable examples include CALAMITI \cite{zuo2021unsupervised}, MURD \cite{liu2024learning}, IGUANe \cite{roca2024iguane}, and STGAN \cite{choi2020stargan}, which have advanced the field by applying deep learning techniques to tackle the challenges of multi-center data harmonization, thus improving the reproducibility and reliability of medical imaging analyses.

\subsection{Contribution and Differences from the Conference Paper}
\label{subsec:conf_paper}
Despite significant advancements, the existing approaches face several challenges. Feature-based methods rely on accurate feature extraction and often assume simplistic statistical relationships, which may fail to capture the inherent complexity of imaging datasets. Image-based methods, while promising, often require large training datasets and struggle to balance visual consistency with the preservation of biologically meaningful information. Moreover, they rely on heavy pre-processing steps and do not generalize to unseen cases.

To overcome the aforementioned issues, in this work we introduce DISARM++, a novel model for harmonizing 3D MR images by addressing inter-scanner variability. DISARM++ disentangles anatomical structure from scanner-specific information to generate \textit{scanner-free} images. This approach preserves the original anatomical structures and biologically informative features, ensuring robust generalizability across different scanners. Our goal is to enable researchers to integrate MRI data from diverse sources without concerns about inconsistencies, while enhancing the extraction of biologically meaningful information. To achieve this, we develop a model that can harmonize images without the need for a new training phase for previously unseen data, allowing seamless integration into various neuroimaging workflows. Unlike traditional preprocessing pipelines, our method retains full-head information without the need for skull-stripping, providing a more comprehensive and less intrusive preprocessing.

The present work is an extended version of the conference paper in \cite{caldera2025disarm}, incorporating several significant improvements: (1) we refined the network architecture and introduced a new loss function, enhancing performance as demonstrated in the ablation study; (2) we trained the model on a larger dataset of MR images; (3) we evaluated the model on a broader range of MR images, including data from healthy individuals, patients with Alzheimer's Disease (AD), and traveling subjects, comparing our approach to state-of-the-art methods; and (4) we conducted several comprehensive downstream analyses to further benchmark our proposal against existing approaches. 

\section{Methodology}
\label{sec:methodology}
In this section, we present the novel proposal designed to harmonize 3D T1-weighted MRI data that extends the DISARM framework introduced in \cite{caldera2025disarm}. The proposed model belongs to the category of I2I translation methods, with its baseline architecture inspired by \cite{lee2020drit++}. The model aims to mitigate batch effects in clinical images acquired from different sources by directly working on the images at the voxel level. Specifically, our approach focuses on the \textit{scanner-free} generation of 3D MRI data with two key objectives: (1) ensuring robust generalizability across a wide range of scanners, including those not seen during training, and (2) eliminating the need for time-consuming preprocessing steps.

\subsection{Mathematical Formulation}
\label{subsec:math_formulation}
Consider a set of MR images \(\mathcal{X} = \bigcup_{i = 1}^{N} \mathcal{X}_i \in \mathbb{R}^{\text{1 x H x W x D}} \), where $\mathcal{X}_i$ represents the collection of images acquired from the \( i \)-th scanner domain among $N$ distinct domains. Since the images are grayscale 3D images, they have dimensions of 1 (channel), H (height), W (width), and D (depth). We assume that the images can be disentangled into two distinct latent spaces $(\mathcal{B}, \mathcal{S})$. Here, $\mathcal{B}$ represents the space that encodes information related to the anatomical structure of the brain, while $\mathcal{S}$ represents the scanner space, which aims to capture information about scanner effects. Thus, an image $\bm{x}$ drawn from $\mathcal{X}$ can be obtained as a combination of $\mathcal{B}$ and $\mathcal{S}$. The \textit{scanner-free} harmonization involves eliminating scanner-specific effects by replacing \(\mathcal{S}\) with random Gaussian noise \(\mathcal{N}(0,1)\), and combining it with \(\mathcal{B}\) into a generator. This process aims to remove scanner-dependent effects while preserving only the anatomical noise-free features encoded in \(\mathcal{B}\). We denote the space in which the images are transferred after \textit{scanner-free} harmonization as \(\mathcal{F}\). 

\subsection{Network Architecture}
\label{subsec:network_architecture}
\begin{figure}[H]
    \centering
    \begin{overpic}[width=1\textwidth]{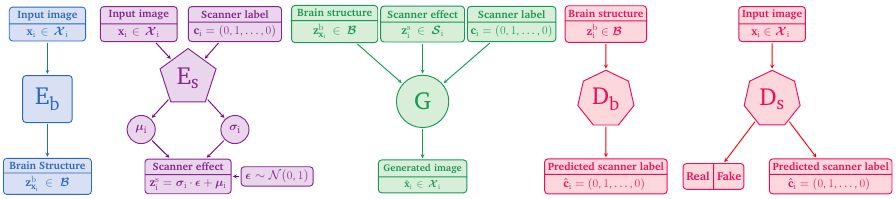}
        \put(3.5,-2.4){\scriptsize (\,a\,)}
        \put(19.2,-2.4){\scriptsize (\,b\,)}
        \put(45.5,-2.4){\scriptsize (\,c\,)}
        \put(66,-2.4){\scriptsize (\,d\,)}
        \put(84.5,-2.4){\scriptsize (\,e\,)}
    \end{overpic}
    \captionsetup{aboveskip=14pt}
    \caption{From left to right, we have a) the brain encoder \(E_b\), b) the scanner encoder \(E_s\), c) the generator \(G\), d) the brain discriminator \(D_b\) and e) the scanner discriminator \(D_s\).}
    \label{fig:modules}
\end{figure}
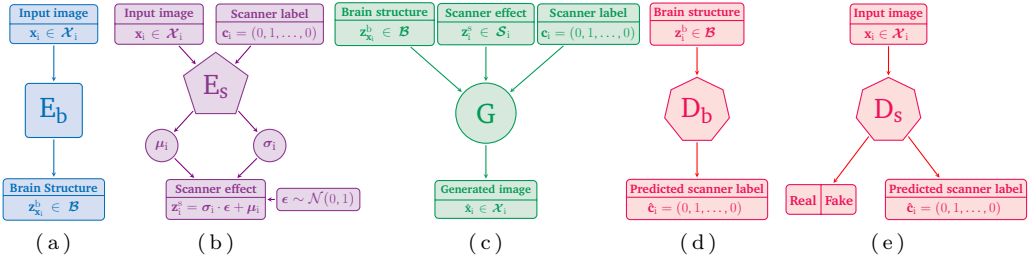

The model architecture consists of five modules, as shown in Figure~\ref{fig:modules}. Let \( \mathcal{C} = \{ \mathbf{c} \in \{0, 1\}^N : \|\mathbf{c}\|_1 = 1 \}\) represent the space of one-hot encoded vectors that describe scanner domains, and \( \mathcal{L} = \{ \mathbf{l} \in \{0, 1\}^2 : \|\mathbf{l}\|_1 = 1 \}\)  represent the space of labels distinguishing real and generated images. The brain encoder (Figure~\ref{fig:modules}a)), $E_b: \mathcal{X} \rightarrow \mathcal{B}$, maps an image $\bm{x} \in \mathcal{X}$ to a lower-dimensional space $\mathcal{B}$, encoding information related to the anatomical structure into a latent vector $\bm{z}_{\bm{x}}^{b}$. The scanner encoder (Figure~\ref{fig:modules}b)), $E_s: (\mathcal{X}, \mathcal{C}) \rightarrow \mathcal{S}$, takes as input an image $\bm{x} \in \mathcal{X}$ and its associated scanner label $\bm{c} \in \mathcal{C}$. Operating as a variational autoencoder, it aims to capture the scanner effect in the image by producing a parametric distribution that models such effect. Specifically, the scanner encoder outputs the mean and variance, characterizing its distribution. For an image drawn from \( \mathcal{X}_i \), which is acquired using the \( i \)-th scanner, the corresponding latent scanner effect vector is denoted as \( \bm{z}_i^s \) and is given by:
\begin{align}
    \bm{z}_i^{s}  = \bm{\sigma}_{i} \cdot \bm{\epsilon} + \bm{\mu}_{i}, \quad \quad \bm{z}_i^{s} \in \mathcal{S}; \quad \bm{\epsilon} \sim \mathcal{N}(0,1).
\end{align}
The generator $G: (\mathcal{B}, \mathcal{S}, \mathcal{C}) \rightarrow \mathcal{X}$ (Figure~\ref{fig:modules}c)) produces an image $\hat{\bm{x}} \in \mathcal{X}$ that preserves a specified brain structure within the space $\mathcal{B}$ while incorporating a scanner attribute from the space $\mathcal{S}$ associated to its label within the space $\mathcal{C}$. The brain discriminator (Figure~\ref{fig:modules}d)) \( D_b: \mathcal{X} \rightarrow \mathcal{C} \) processes an image \( \bm{x} \in \mathcal{X} \) and aims to predict the scanner label \( \bm{c} \in \mathcal{C} \), indicating the scanner used to acquire the image. Finally, the scanner discriminator \( D_s: \mathcal{X} \rightarrow (\mathcal{L}, \mathcal{C}) \) (Figure~\ref{fig:modules}e)) takes an image \( \bm{x} \in \mathcal{X} \) or \( \hat{\bm{x}} \in \mathcal{X} \) as input and attempts to determine whether the image is real or generated by the generator, as well as its associated scanner label \( \bm{c} \in \mathcal{C} \).
Unlike the baseline DISARM model, the proposed approach processes thinner 3D volumes consisting of 26 slices-wide moving window rather than full 3D image volumes of the MR images. Volumes are then merged at inference time. This modification reduces computational complexity and memory requirements, allowing the inclusion of more sophisticated model layers that enhance feature extraction capabilities while maintaining the ability to learn meaningful representations from spatial contexts. 
A further key innovation is the integration of channel and spatial attention layers into the modules \(E_b\), \(G\), and \(D_b\). This attention mechanism is crucial for highlighting significant features, thereby enhancing the model's ability to embed anatomical structures. This ensures that critical structural details are better preserved during the generation of harmonized images, leading to higher fidelity in the final output. In addition, a novel loss function is introduced compared to the baseline model, to improve the \textit{scanner-free} space.
 
\subsection{Training Process}
\label{subsec:train_process}
\begin{figure}[H]
    \centering
    \includegraphics[width=1\textwidth]{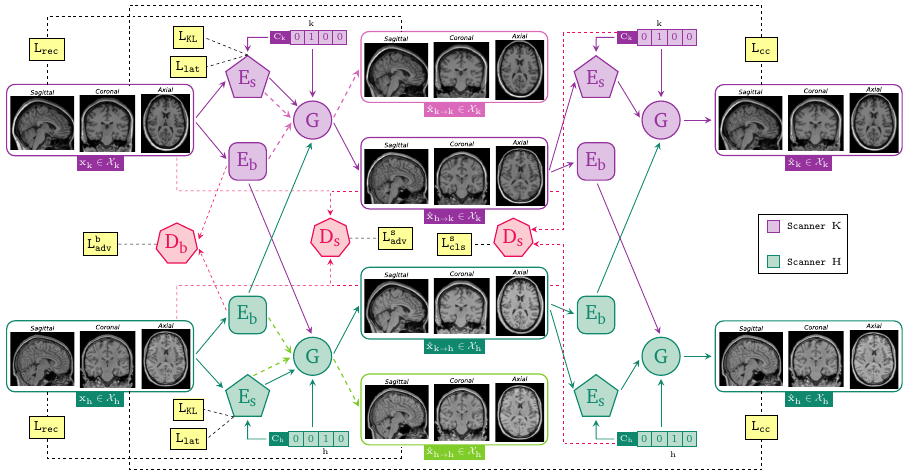}
    \caption{A high-level functional diagram of the model training procedure. The encoders and the generator are single modules, despite being depicted in different colors. The varying colors for the encoders highlight the specific acquisition scanner domain of the images they process. For the generator, the different colors emphasize the scanner domain to which it is transferring the image. The images used to illustrate the procedure are acquired with the \texttt{Prisma} (purple) and \texttt{Gyroscan Intera} (green) scanners.}
    \label{fig:train_procedure}
\end{figure}
\begin{figure}[H]
    \centering
    \includegraphics[width=1\textwidth]{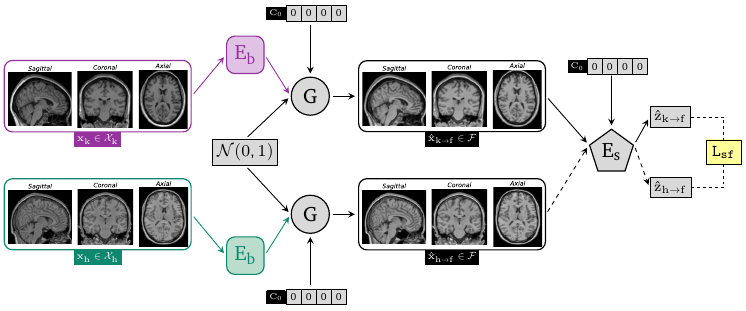}
    \caption{Training procedure part related to the \textit{scanner-free} loss. The images used to illustrate the procedure are acquired with the \texttt{Prisma} (purple) and \texttt{Gyroscan Intera} (green) scanners.}
    \label{fig:scanner_free_loss}
\end{figure}

In this section, we outline the training procedure referring to Figure~\ref{fig:train_procedure} and Figure~\ref{fig:scanner_free_loss}.
During each iteration, the training procedure involves randomly selecting two scanner domains from the pool of $N$ domains. For clarity, we describe the procedure assuming that two images have been sampled: one from the scanner domain $K$ (represented in purple in Figure~\ref{fig:train_procedure}), denoted by the pair $(\bm{x}_k, \bm{c}_k)$, and another from scanner domain $H$ (represented in green in Figure~\ref{fig:train_procedure}), denoted by the pair $(\bm{x}_h, \bm{c}_h)$. 

In the initial step, for each image \( \bm{x}_i \) (where \( i = \{h, k\} \)), we extract the anatomical structure embedding \( \bm{z}_{\bm{x}_i}^b = E_b(\bm{x}_i) \) using the brain encoder \( E_b \), and the scanner effect embedding \( \bm{z}_i^s = E_s(\bm{x}_i) \) using the scanner encoder \( E_s \).
Note that we denote the scanner effect embedding as $\bm{z}_i^s$ rather than $\bm{z}_{x_i}^s$ because it represents the characteristics of the scanner-specific image family as a whole, rather than the noise embedded in an individual image.
At this point, the anatomical embeddings \( \bm{z}_{\bm{x}_k}^b \) and \( \bm{z}_{\bm{x}_h}^b \) are swapped and used as inputs to the generator. The generator then synthesizes two new images: \( \hat{\bm{x}}_{h \rightarrow k} = G(\bm{z}_{\bm{x}_h}^b, \bm{z}_k^s, \bm{c}_k) \), which combines the anatomical structure of \( \bm{x}_h \) with the scanner effect of \( \bm{x}_k \), and \( \hat{\bm{x}}_{k \rightarrow h} = G(\bm{z}_{\bm{x}_k}^b, \bm{z}_h^s, \bm{c}_h) \), which integrates the scanner effect of \( \bm{x}_h \) with the anatomical structure of \( \bm{x}_k \). Thereafter, the process of swapping anatomical structure embeddings is repeated—this time for the newly generated images \( \hat{\bm{x}}_{h \rightarrow k} \) and \( \hat{\bm{x}}_{k \rightarrow h} \)—resulting in the cyclic reconstruction of the input images as \( \hat{\bm{x}}_k = G(\bm{z}_{\hat{\bm{x}}_{k \rightarrow h}}^b, \bm{z}_{h \rightarrow k}^s, \bm{c}_k) \) and \( \hat{\bm{x}}_h = G(\bm{z}_{\hat{\bm{x}}_{h \rightarrow k}}^b, \bm{z}_{k \rightarrow h}^s, \bm{c}_h) \). The generator is also employed to generate \( \hat{\bm{x}}_{i \rightarrow i} = G(\bm{z}_{\bm{x}_i}^b, \bm{z}_i^s, \bm{c}_i) \) (where \( i = \{h, k\} \)) by combining the latent representations extracted from the same image, enabling the direct reconstruction of the input images.
Moreover, as shown in Figure~\ref{fig:scanner_free_loss}, during each iteration, the anatomical embeddings \( \bm{z}_{\bm{x}_k}^b \) and \( \bm{z}_{\bm{x}_h}^b \) for both images are fed into the generator, along with the same random noise \( \mathcal{N}(0,1) \). This enables the \textit{scanner-free} generation of two new images, \( \hat{\bm{x}}_{i \rightarrow f} = G(\bm{z}_{\bm{x}_i}^b, \bm{\epsilon}, \bm{c}_0) \) (where \( i = \{h, k\} \)). All these reconstructions enable the formulation of various loss functions, each addressing different aspects of the model, which will be detailed in Section~\ref{subsec:losses}.

\subsection{Loss Functions}
\label{subsec:losses}
In this section, we define the objective function of the model whose minimization guides the optimization process. By minimizing this function during each iteration of the training procedure, we can iteratively adjust and refine the parameters of the network modules.

\subsubsection*{Cycle Consistency Loss \textnormal{($L_{\text{cc}}$)} } 
As the proposed model follows a cycle-GAN architecture and training process, the first loss we introduce ensures cross-cycle consistency within the process described in Figure~\ref{fig:train_procedure}. To do this, we enforce the similarity between the original images $\bm{x}_k$ and $\bm{x}_h$ and their reconstructed versions $\hat{\bm{x}}_k$ and $\hat{\bm{x}}_h$:

\footnotesize
\begin{align}
L_{cc} = \mathbb{E}_{\bm{x}_k,\bm{x}_h} \bigg{[} \ \big{\lVert} \hat{\bm{x}}_k - \bm{x}_k \big{\rVert} +  \big{\lVert} \hat{\bm{x}}_h  - \bm{x}_h  \big{\rVert} \bigg{]}. 
\end{align}
\normalsize

\subsubsection*{Self Reconstruction Loss \textnormal{($L_{\text{rec}}$)} } 
Similar to the previous loss, we seek to ensure that the input images, $\bm{x}_k$ and $\bm{x}_h$, closely resemble their directly reconstructed counterparts, \( \hat{\bm{x}}_{k \rightarrow k} \) and \( \hat{\bm{x}}_{h \rightarrow h} \):

\footnotesize
\begin{align}
L_{\text{rec}} = \mathbb{E}_{\bm{x}_k, \bm{x}_h} \bigg{[} \ \big{\lVert} \hat{\bm{x}}_{k \rightarrow k} - \bm{x}_k \big{\rVert} + \big{\lVert} \hat{\bm{x}}_{h \rightarrow h} - \bm{x}_h \big{\rVert} \bigg{]} .
\end{align}
\normalsize

\subsubsection*{Brain Structure Adversarial Loss \textnormal{(\( L_{\text{adv}}^b \))} } 
In order to produce scanner-independent anatomical structure embeddings, we aim at mapping them into a shared space \( \mathcal{B} \) where domain membership is indistinguishable. To achieve this, adversarial training for the brain encoder \( E_b \) is employed. Specifically, the latent representations of the brain structure \( \bm{z}_{\bm{x}_k}^b  \) and \( \bm{z}_{\bm{x}_h}^b  \) are input to the discriminator \( D_b \) (Figure~\ref{fig:train_procedure}), which learns to discriminate between their scanner memberships. Meanwhile, the encoder \( E_b \) learns to produce anatomical structure embeddings that are indistinguishable in terms of the membership of the scanner domain by \( D_b \). The loss function is formally defined as

\footnotesize
\begin{align}
\label{eq: brain struct loss}
L^{\text{b}}_{\text{adv}} = \frac{1}{2} \, \mathbb{E}_{\bm{x}_k} \bigg{[}  \log \Bigl[ D_b(\bm{z}_{\bm{x}_k}^b) ( 1 - D_b(\bm{z}_{\bm{x}_k}^b ) ) \Bigr] \bigg{]} \ + \frac{1}{2} \, \mathbb{E}_{\bm{x}_h} \bigg{[} \log \Bigl[ D_b(\bm{z}_{\bm{x}_h}^b) ( 1 - D_b(\bm{z}_{\bm{x}_h}^b) ) \Bigr] \bigg{]}.
\end{align}
\normalsize

\subsubsection*{Scanner Classification Loss \textnormal{($L^{\text{s}}_{\text{cls}}$)} }
To force the generator \( G \) to adopt the desired scanner-related information, i.e., the one plugged into it, during the generation of a new image, the discriminator $D_s$ is trained to predict the scanner label of the generated images \( \hat{\bm{x}}_{h \rightarrow k} \) and \( \hat{\bm{x}}_{k \rightarrow h} \) as follows

\footnotesize
\begin{align}
L^{\text{s}}_{\text{cls}} = \mathbb{E}_{\bm{x}_k} \bigg{[} - \log\big{[}D_c(\bm{c}_k | \hat{\bm{x}}_{h \rightarrow k} )\big{]} \bigg{]} \ + \ \mathbb{E}_{\bm{x}_h} \bigg{[} - \log\big{[} D_c(\bm{c}_h | \hat{\bm{x}}_{k \rightarrow h} )\big{]} \bigg{]} .
\end{align}
\normalsize

\subsubsection*{Scanner Adversarial Loss \textnormal{($L^{\text{s}}_{\text{adv}}$)} } 
Similarly, with the goal of training the generator \( G \) to produce realistic images in each specific scanner domain \( \mathcal{X}_i \) through adversarial training as illustrated in Figure~\ref{fig:train_procedure}, the discriminator \( D_s \) receives both real images \( \bm{x}_k \) and \( \bm{x}_h \) and generated images \( \hat{\bm{x}}_{h \rightarrow k} \) and \( \hat{\bm{x}}_{k \rightarrow h} \). It then attempts to discriminate between real and generated images within their respective scanner domains \( \mathcal{X}_k \) and \( \mathcal{X}_h \), as follows

\footnotesize
\begin{align}
L^{\text{s}}_{\text{adv}} &= \frac{1}{2} \, \mathbb{E}_{\bm{x}_k} \bigg{[}  \log \Bigl[ D_s(\bm{x}_k) ( 1 - D_s(\hat{\bm{x}}_{h \rightarrow k}) ) \Bigr] \bigg{]} \, + \, \frac{1}{2} \, \mathbb{E}_{\bm{x}_h} \bigg{[}  \log \Bigl[ D_s(\bm{x}_h) ( 1 - D_s(\hat{\bm{x}}_{k \rightarrow h}) ) \Bigr]  \bigg{]} 
\end{align}
\normalsize

\subsubsection*{Scanner-Free Loss \textnormal{($L_{\text{sf}}$)} }
We introduce a novel loss specifically designed to ensure that the generator consistently reproduces the same scanner effect when provided with identical random noise inputs. To achieve this, \( \hat{\bm{x}}_{k \rightarrow f} \) and \( \hat{\bm{x}}_{h \rightarrow f} \), generated as described in Section~\ref{subsec:train_process}, are fed to the scanner encoder \(E_s\) along with a null vector \(\bm{c}_0\). The encoder extracts their respective latent representations of the scanner effect, \( \hat{\bm{z}}_{k \rightarrow f} \) and \( \hat{\bm{z}}_{h \rightarrow f} \), and the loss enforces these representations to be as similar as possible (Figure~\ref{fig:scanner_free_loss}):

\footnotesize
\begin{align}
L_{\text{sf}} &= \mathbb{E}_{\bm{x}_k,\bm{x}_h} \bigg{[}  \big{\lVert} E_s\big{(} \hat{\bm{x}}_{k \rightarrow f} \big{)} - E_s\big{(} \hat{\bm{x}}_{h \rightarrow f}  \big{)}  \big{\rVert} \bigg{]} .
\end{align}
\normalsize

\subsubsection*{Total Objective Function} 

The overall model loss is defined as follows
\footnotesize
\begin{align}
L_{\text{tot}} &= \lambda_{\text{cc}} L_{\text{cc}} +  \lambda_{\text{rec}} L_{\text{rec}} + \lambda_{\text{lat}} L_{\text{lat}} + \lambda_{\text{KL}} L_{\text{KL}} + \lambda_{\text{sf}} L_{\text{sf}}  - \lambda_{\text{adv}}^{\text{b}} L^{\text{b}}_{\text{adv}} - \lambda_{\text{cls}}^{\text{s}} L^{\text{s}}_{\text{cls}} - \lambda_{\text{adv}}^{\text{s}} L^{\text{s}}_{\text{adv}} 
\end{align}
\normalsize
where, the Kullback–Leibler divergence loss $L_{\text{KL}}$ focuses on aligning the scanner effect embeddings with a standard Gaussian prior and $L_{\text{lat}}$ ensures that the mean vector $\bm{\mu}_i$, produced by the scanner encoder $E_s$, remains close to a standard Gaussian distribution.

\subsection{Inference}
\label{subsec:inference}
\begin{figure}[H]
    \centering
    \includegraphics[width=1\textwidth]{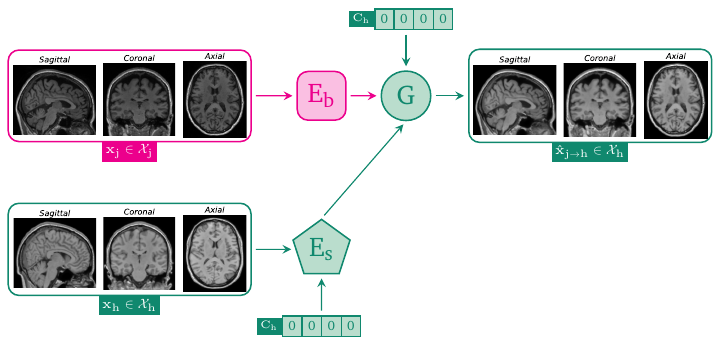}
    \caption{Inference to a reference scanner for a new image \(\bm{x}_{\text{j}} \in \mathcal{X}_{\text{j}}\). Specifically, the figure illustrates the transfer of an image acquired with the \texttt{Skyra Fit} (pink) scanner to the reference \texttt{Gyroscan Intera} (green).}
    \label{fig:reference_inference}
\end{figure}

\begin{figure}[H]
    \centering
    \includegraphics[width=1\textwidth]{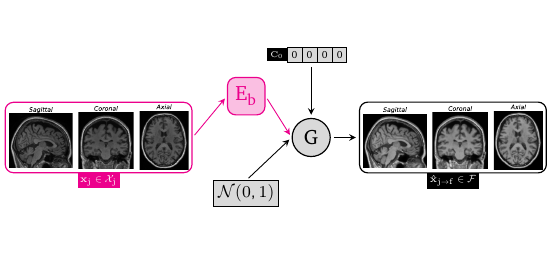}
    \caption{Inference to \textit{scanner-free} for a new image \(\bm{x}_{\text{j}} \in \mathcal{X}_{\text{j}}\). Specifically, the figure illustrates the transfer of an image acquired with the \texttt{Skyra Fit} (pink) scanner to the \textit{scanner-free} space (black).}
    \label{fig:scanner_free_inference}
\end{figure}

Consider a new image, \(\bm{x}_{\text{j}} \in \mathcal{X}_{\text{j}}\), which may be acquired either from one of the training scanners (\( j \in \{1, \hdots, N\} \)) or from a previously unseen scanner not included in the pool of training scanners (\( j \notin \{1, \hdots, N\} \)). During inference, the image can be harmonized in two distinct ways. One approach involves transferring the new image into the space of one of the training scanners, using it as a reference. To achieve this, we employ the brain encoder \(E_b\) to extract the anatomical embedding of the new image \( \bm{z}_{\bm{x}_j}^b \). This representation is then combined, using the generator \(G\), with the scanner effect \( \bm{z}_i^s \), where \( i \in \{1, \hdots, N\} \), corresponding to one of the training scanners (Figure~\ref{fig:reference_inference}). By harmonizing all desired images into the same scanner space, this method ensures a uniform scanner effect across the harmonized dataset. The second approach involves transferring the new image into the new \textit{scanner-free} space. Similar to the first method, we use the brain encoder \( E_b \) to extract the anatomical embedding \( \bm{z}_{\bm{x}_j}^b \). However, instead of plugging a scanner effect, we use a random Gaussian noise \( \mathcal{N}(0,1) \) to be fed into the generator \( G \) together with the brain information (Figure~\ref{fig:scanner_free_inference}). This approach effectively harmonizes all desired images into a shared, scanner-independent space, ensuring uniformity across the dataset without introducing scanner-specific variations.
This configuration behaves like a denoising step.

\section{Datasets and Preprocessing}
\label{sec:datasets_and_prep}

\subsection{Datasets}
\label{subsec:datasets}
We obtained T1-weighted MR images of healthy controls from five distinct datasets. Specifically, the collection includes 313 images from the Alzheimer's Disease Neuroimaging Initiative (ADNI3) \cite{jack2008alzheimer} acquired using five different scanners, 60 images from the Parkinson's Progression Markers Initiative (PPMI) \cite{marek2011parkinson} obtained with three scanners, 581 images from the IXI Brain Development Dataset (IXI) \cite{ixidata} acquired with three scanners, 494 images from the Southwest University Adult Lifespan Dataset (SALD) \cite{wei2017structural} captured using a single scanner, and 117 images from a private dataset provided by the Italian Neuroimaging Network (RIN) \cite{nigri2022quantitative}, collected with six different scanners. We also considered patients with dementia exhibiting atrophy, including 41 MR scans from the private NeuroArtP3 \cite{malaguti2024artificial} dataset and 61 MR scans from the public ADNI3 dataset of AD patients. In addition, we included seven subjects from the Strategic Research Program for Brain Science (SRPBS) traveling subjects dataset \cite{tanaka2021multi}, with each subject having between 8 and 11 MR scans collected at different sites. Detailed descriptions of the datasets — including scanner types, manufacturers, number of images, field strengths, and participants' age — are provided in Appendix A of the Supplementary Materials.

\subsection{Preprocessing Steps}
\label{subsec:preprocessing}
Image preprocessing was conducted using the FSL library \citep{fsl1, fsl2}. Initially, the images were standardized to a common orientation. Next, bias-field correction was applied to address magnetic field variations in the MRI scanner, which can lead to blurring, loss of detail, and altered pixel intensities. The \texttt{FAST} algorithm within FSL was used for this correction, combining the non-parametric N3 bias field correction algorithm with a parametric model-based approach. The images were then registered to a standard space using the \texttt{FLIRT} command in FSL, which employs an affine transformation model to align the images through translations, rotations, and scaling, mapping them to the Standard MNI152-T1-1mm space. After preprocessing, the resulting images had dimensions of $(1, \ 182, \ 218, \ 182)$.

\section{Experiments}
\label{sec:experiments}
We validated our approach using T1-weighted MR images from healthy controls. Specifically, we trained DISARM++ on a dataset comprising 701 images from the ADNI3~\cite{jack2008alzheimer}, PPMI~\cite{marek2011parkinson}, and IXI~\cite{ixidata} datasets (Table~\ref{tab:train_dataset}). These training images were acquired using $N=5$ different scanner models: 167 images with Prisma Fit, 69 with Prisma, 30 with Achieva dStream, 250 with Gyroscan Intera, and 185 with Intera. To improve the robustness of the model, we used a data augmentation procedure through dense random elastic deformation \cite{torchioLibrary}. Additional details about the training procedure are provided in Appendix B.

\begin{table}[H]
\footnotesize
\scriptsize
\centering
\begin{tblr}{@{} >{\raggedright\arraybackslash} p{1.5cm} >{\raggedright\arraybackslash}p{3.0cm} >{\raggedright\arraybackslash}p{2.2cm} >{\raggedright\arraybackslash}p{1.8cm} >{\raggedright\arraybackslash}p{1.3cm} >{\centering\arraybackslash}p{1.2cm} @{}}
\hline[0.5pt]
\textbf{}  & \textbf{Scanner Model} & \textbf{Manufacturer} & \textbf{Dataset} & \textbf{Img. \#} & \textbf{Total} \\ 
\hline \hline
\SetCell[r=6]{c,3.0cm} \textbf{Training} & Prisma Fit & SIEMENS & ADNI3  & 167/167 & \SetCell[r=6]{c,1.2cm} 701 \\
&  Prisma & SIEMENS & ADNI3  & 69/69   \\
&  Achieva dStream & Philips & ADNI3  & 26/26   \\
&  Achieva dStream  & Philips & PPMI  & 5/5   \\
&  Gyroscan Intera & Philips & IXI & 250/322 \\
&  Intera & Philips & IXI & 185/185 \\
\hline[0.5pt]
\end{tblr}
\caption{Description of the healthy controls training datasets, including details on scanner types, manufacturers, source datasets, and number of images. Values in the "Img. \#" column represent the number of images used out of the total available for that scanner and dataset combination.}
\label{tab:train_dataset}
\end{table}

In these sections, we first describe the evaluation procedures used in our experiments (Section \ref{subsec:metrics_and_stat_tets}), followed by the ablation study (Section \ref{subsec:ablation_study}), a comparison with state-of-the-art methods (Section \ref{subsec:soa_comparison}), and downstream analyses (Section \ref{subsec:down_analysis}).

\subsection{Evaluation Metrics and Statistical Tests}
\label{subsec:metrics_and_stat_tets}
In this section, we detail the metrics used in our evaluations. For a more in-depth explanation of these metrics and their computation within our specific context, refer to Appendix C of the Supplementary Materials.

\subsubsection{Anatomical Structure Metrics}
\label{subsubsec:struct_metrics}
To evaluate the preservation of anatomical structures and the quality of generated images (see Table~\ref{tab:metrics_descr}), we employed four key metrics: (1) the structural component of the Structural Similarity Index Measure (SSIM) \cite{wang2004image}, known as Struct-SSIM, (2) the complete SSIM index \cite{wang2004image}, (3) the Learned Perceptual Image Patch Similarity (LPIPS) metric \cite{zhang2018unreasonable}, and (4) the Fréchet Inception Distance (FID) \cite{heusel2017gans}. The SSIM metrics and LPIPS are computed between image pairs, whereas FID compares two sets of images. The complete SSIM index assesses similarity based on luminance, contrast, and structure, with higher scores indicating greater overall similarity. In contrast, Struct-SSIM focuses specifically on structural similarity by evaluating local spatial patterns independently of luminance and contrast, thereby capturing fine details and spatial coherence; higher Struct-SSIM values signify better structural similarity. LPIPS measures perceptual similarity by comparing feature activation maps from a deep neural network, which makes it sensitive to semantic and textural differences that align with human visual perception; lower LPIPS scores indicate higher similarity. Finally, FID evaluates the distributional similarity between real and generated images by comparing feature embeddings extracted from a pre-trained Inception network. It effectively assesses the overall image quality and realism, with lower values reflecting closer alignment to the real data distribution. Notably, FID, LPIPS, or complete SSIM consider not only structural differences but also luminance and contrast, making them sensitive to the extent of harmonization performed by different models. As a result, using these indices would make it difficult to compare the preservation of anatomical structure across different models. We use FID and LPIPS in the ablation study (Section \ref{subsec:results_ablation_study}), where the level of harmonization across different DISARM++ configurations remains comparable, and the complete SSIM in the traveling subjects evaluation (Section \ref{subsubsec:travel_sbj}), where images similarity can be assessed in terms of contrast, luminance, and structure.

\subsubsection{Harmonization Metrics}
\label{subsubsec:harm_metrics}
To evaluate the transfer of scanner characteristics, we assessed the similarity of voxel intensity distributions before and after harmonization using three metrics: Jensen-Shannon Divergence (JSD) \cite{lin1991divergence}, Hellinger Distance (HD) \cite{kailath2003divergence, rao1995review}, and Wasserstein Distance (WD) \cite{villani2008optimal} (see Table~\ref{tab:metrics_descr}). Practically, given the MRI scans having dimension \( (1, H, W, D) \) — where \( H \) represents height, \( W \) represents width, and \( D \) represents depth — we computed these metrics based on the set of $1 \times H \times W \times D$ voxel intensity distributions derived from MRI scans, considering pre- and post-harmonization data. Specifically, we compute the empirical distribution by estimating the underlying probability distribution of the voxel intensities for each unfolded image, and we average the distributions of images belonging to the same scanner. In this way, we obtain  \( G \) distributions — with \( G \) being the number of the test scanners — from pre-harmonization images and \( G \) distributions from post-harmonization images. In each of the two sets of $G$ distributions, we computed the values for the three metrics across all possible pairs to quantify the similarity between them. Therefore, when reporting these metrics, we present the mean and standard deviation of all pairwise comparison values, providing a general assessment for all scanners.  We perform this step for both pre-harmonization and post-harmonization sets, enabling the comparison of the similarity between the distributions before and after harmonization. To statistically evaluate the effectiveness of harmonization, we perform a paired t-test comparing the values of the similarity metrics before and after harmonization. This test assesses whether the mean difference is significantly different from zero. If the differences do not follow a normal distribution, we employ a bootstrap paired t-test, resampling the data to estimate the distribution of differences. We report the 95\% confidence interval (CI) for these differences; if the CI does not include zero, it indicates a significant effect of harmonization on voxel intensity similarity.

Besides the three aforementioned indexes, in certain analyses, we also employed the K-sample Anderson-Darling test (AD-test) \cite{Anderson-Darling}, a non-parametric test that evaluates whether multiple samples originate from the same distribution. This test was applied to the set of \( G \) distributions, separately for pre- and post-harmonization images. If we accept the null hypothesis, it suggests no significant difference between distributions, indicating successful harmonization. Conversely, rejecting the null hypothesis implies that at least one distribution differs, suggesting incomplete harmonization.

\begin{table}[H]
\tiny
\centering
\begin{tblr}{@{} >{\raggedright\arraybackslash} m{0.9cm}  >{\centering\arraybackslash} m{2.1cm} >{\centering\arraybackslash} m{0.8cm} >{\centering\arraybackslash} m{1.7cm} >{\centering\arraybackslash} m{6.7cm} @{}}
\hline[0.5pt]
 \textbf{Metric} & \textbf{Description} & \textbf{Range} & \textbf{Interpretation} & \textbf{Formula} \\
\hline \hline
\textbf{FID} & \textit{Measures the distribution distance between real and generated image features.} & [0, $\infty$] & \textit{10–20\\ (Good) 50+\\ (Poor)} &  $\text{FID} = \| \mu_r - \mu_g \|^2 + \text{Tr}\bigg( \Sigma_r + \Sigma_g - 2\sqrt{\Sigma_r\Sigma_g} \bigg)$  \\
\hline
\textbf{LPIPS} & \textit{Measures perceptual similarity between images based on learned deep features.} & [0,1] & \textit{0.1–0.3\\ (Good) 0.5+\\ (Poor)} & $\text{LPIPS}(x, y) = \sum_l \frac{1}{H_l W_l} \sum_{h,w} \| w_l \big( f^l(x)_{hw} - f^l(y)_{hw} \big) \|^2$ \\
\hline
\textbf{Struct-SSIM}  & \textit{Measures the structural similarity between two images.} & [-1,1] & \textit{Higher values, Higher similarity} & $\text{Struct-SSIM}(x, y) = \frac{\sigma_{xy} + C_3}{\sigma_x \sigma_y + C_3}$ \\
\hline
\textbf{SSIM}  & \textit{Measures the similarity considering luminance, contrast, and structure.} & [-1,1] & \textit{Higher values, Higher similarity} & $\text{SSIM}(x, y) = \frac{(2\mu_x\mu_y + C_1)(2\sigma_{xy} + C_2)}{(\mu_x^2 + \mu_y^2 + C_1)(\sigma_x^2 + \sigma_y^2 + C_2)}$ \\
\hline
\end{tblr}
\vspace{1mm} \\
{\text{\scriptsize (\,a\,)}}
\vspace{3mm} \\
\centering
\begin{tblr}{@{} >{\raggedright\arraybackslash} m{1.0cm}  >{\centering\arraybackslash} m{2.8cm} >{\centering\arraybackslash} m{1.2cm} >{\centering\arraybackslash} m{2.2cm} >{\centering\arraybackslash} m{5cm} @{}}
\hline[0.5pt]
 \textbf{Metric} & \textbf{Description} & \textbf{Range} & \textbf{Interpretation} & \textbf{Formula} \\
\hline \hline
\textbf{JSD} & \textit{Measures the similarity between two probability distributions based on KL divergence.} & [0, $\log(2)$] & \textit{Lower values, Higher similarity} & $\text{JSD}(P, Q) = \frac{1}{2} \text{KL}(P \| M) + \frac{1}{2} \text{KL}(Q \| M)$ \\
\hline
\textbf{HD} & \textit{Measures the similarity between two probability distributions using their square root.} & [0, 1] & \textit{Lower values, Higher similarity} & $\text{HD}(P, Q) = \frac{1}{\sqrt{2}} \sqrt{\sum_{x} \left( \sqrt{P(x)} - \sqrt{Q(x)} \right)^2}$ \\
\hline
\textbf{WD} & \textit{Measures the minimum cost of transporting mass to transform one distribution into another.} & [0, $\infty$] & \textit{Lower values, Better match} & $\text{WD}(P, Q) = \int_{\mathbb{R}} \left| F_P(x) - F_Q(x) \right| dx$ \\
\hline
\end{tblr}
\vspace{1mm} \\
{\text{\scriptsize (\,b\,)}}
\caption{Evaluation metrics. a) Anatomical structure metrics: For FID \cite{heusel2017gans}, \(\mu_r\) and \(\mu_g\) represent the means, and \(\Sigma_r\) and \(\Sigma_g\) are the covariances of the real and generated image features. For LPIPS \cite{zhang2018unreasonable}, \(f^l(x)\) and \(f^l(y)\) are the deep features at layer \(l\) for images \(x\) and \(y\), \(w_l\) are the learned weights, and \(H_l\) and \(W_l\) denote the height and width of the feature map at layer \(l\). For SSIM (Structural) \cite{wang2004image}, \(\sigma_{xy}\) is the covariance between \(x\) and \(y\), \(\sigma_x\) and \(\sigma_y\) are their standard deviations, and \(C_3\) is a stabilization constant. For SSIM (Complete), \(\mu_x\) and \(\mu_y\) are the means, \(\sigma_x^2\) and \(\sigma_y^2\) are the variances, \(\sigma_{xy}\) is the covariance, and \(C_1\) and \(C_2\) are constants. 
b) Harmonization metrics: For Jensen-Shannon Divergence (JSD) \cite{lin1991divergence}, Hellinger Distance (HD) \cite{nikulin2001hellinger}, and Wasserstein Distance (WD)  \cite{panaretos2019statistical}, \(P\) and \(Q\) are the probability distributions being compared, and \(x\) denotes a specific outcome in the probability space. \(M = \frac{1}{2}(P + Q)\) is the mixed distribution, and \(\text{KL}(P \| M)\) represents the Kullback-Leibler divergence between \(P\) and \(M\). \(F_P(x)\) and \(F_Q(x)\) denote the cumulative distribution functions of \(P\) and \(Q\), respectively.}
\label{tab:metrics_descr}
\end{table}

\subsection{Ablation Study}
\label{subsec:ablation_study}
In this section, we present the ablation study of the proposed model by systematically removing or modifying specific components. We begin by evaluating the performance of the baseline DISARM model \cite{caldera2025disarm}. Next, starting with the complete model, we analyze the effects of removing key components: the \textit{scanner-free} loss ($L^{\text{sf}}$), the KL divergence loss ($L^{KL}$), the latent loss ($L^{lat}$), and the attention layers. The training for all configurations was conducted using 701 images from six different scanners, as described in Section~\ref{sec:experiments}. The evaluation was conducted by harmonizing the 250 test images from five different scanners detailed in Table~\ref{tab:ablation_dataset} into the \textit{scanner-free} space across all model configurations. Here we compare the different configurations in terms of preservation of anatomical structure and image quality using Struct-SSIM, LPIPS, and FID metrics. To assess harmonization performance, we used the JSD and the AD-test \cite{Anderson-Darling}, considering the mean voxel intensity distributions from the five test scanners. The results of the ablation study are presented in Section \ref{subsec:results_ablation_study}.

\begin{table}[H]
\footnotesize
\scriptsize
\centering
\begin{tblr}{@{} >{\raggedright\arraybackslash} p{2.5cm} >{\raggedright\arraybackslash}p{2.5cm} >{\raggedright\arraybackslash}p{1.8cm} >{\centering\arraybackslash}p{1.8cm} >{\centering\arraybackslash}p{1.8cm}  @{}}
\hline[0.5pt]
\textbf{Evaluation type}  & \textbf{Scanner Model} & \textbf{Manufacturer} & \textbf{Dataset} & \textbf{Images No.} \\ 
\hline \hline
\SetCell[r=5]{c,3.5cm} \textbf{Ablation study} & Prisma & SIEMENS & RIN  & 40/40  \\
&  Triotim & SIEMENS & PPMI & 41/41   \\
&  Gyroscan Intera & Philips & IXI & 72/322 \\
&  Unknown & GE & IXI & 74/74 \\
&  Ingenia CX & Philips & RIN & 23/23  \\
\hline[0.5pt]
\end{tblr}
\caption{Description of the test images utilized in the ablation study, including details on scanner types, manufacturers, source datasets, and number of images.}
\label{tab:ablation_dataset}
\end{table}

\subsection{Comparison with State-of-the-Art}
\label{subsec:soa_comparison}
This section outlines the evaluation of the harmonization results produced by the proposed model, comparing them to benchmark methods (Section \ref{subsubsec:benchmark_methods}). We detail the datasets used for each evaluation and describe the evaluation process. The results are then provided in Section \ref{subsec:results_harmonization}.

\subsubsection{Benchmarking Methods}
\label{subsubsec:benchmark_methods}
To benchmark our model, we compare the performance of the proposed model with the following image-based approaches:
\begin{itemize}
    \item \textbf{STGAN} \cite{choi2020stargan}: We utilize the pre-trained STGAN model, which was trained on 718 images from five different dataset subsets: ADNI3 (42 images), ICBM (200 images), UKBB (200 images), PPMI (76 images), and ABCD (200 images). The preprocessing steps involved skull-stripping the images \cite{segonne2004hybrid} and correcting for nonuniformity using the N3 method \cite{sled1998nonparametric} in Freesurfer. Additionally, the images were linearly registered to the standard MNI template (ICBM 152 Nonlinear Symmetric atlas \cite{fonov2009unbiased}) and resized to isotropic 1 $mm^3$ voxels. Following the STGAN implementation, harmonization was performed on a sliding window of three image slices with a stride of 1. The final T1 harmonized MRI 3D volumes were reconstructed by combining these harmonized partial volumes. The model assumes that each image belongs to a unique domain and can be decomposed into its content and style. Therefore, for comparison with other models, we harmonize the test images by using an image from the \texttt{Gyroscan Intera} scanner as a reference. 
     \item \textbf{IGUANe} \cite{roca2024iguane}: We utilize the IGUANe model, trained on a dataset of 4,347 T1-weighted brain MRI images from 11 distinct scanners across eight public studies: SALD, IXI, OASIS-3, NKI-RS, NMorphCH, AIBL, HCP, and ICBM. Preprocessing involved skull-stripping using HD-BET \cite{isensee2019automated}, bias field correction with N4ITK \cite{tustison2010n4itk}, linear registration to the MNI 1 mm³ space using FSL-FLIRT \cite{jenkinson2002improved}, cropping to 160×192×160 voxels, and standardizing intensities by dividing by the median brain intensity. Images were then scaled so that the median brain intensity matched a value of 1 while maintaining a background intensity of 0. The IGUANe model, based on a 3D extension of the CycleGAN framework, implements a many-to-one adversarial training strategy to harmonize MRI data from various acquisition sites into a common reference domain, represented by the SALD dataset. The authors selected the SALD dataset as a reference because it contains a large number of MR images and covers a wide age range.
\end{itemize}

\subsubsection{Evaluation Setup}
\label{subsubsec:evaluation_setup}
To evaluate the harmonization results obtained with the proposed model in comparison to benchmark methods, we use the results from the set detailed in Table~\ref{tab:test_dataset}, which included 796 healthy control MR images from the RIN, IXI, SALD and PPMI datasets acquired from 10 different scanners, along with the SRPBS traveling subject dataset.

The first analysis focused on visualizing the harmonization results by presenting slices from each direction (axial, coronal, and sagittal). For each model, we display 10 original images (one from each test scanner) alongside their harmonized counterparts. Additionally, we provide heatmaps illustrating pixel-wise differences between the harmonized images and their originals. For the proposed model, we report the visualization of the transfer of images to the \textit{scanner-free} space. The results are presented in Section~\ref{subsubsec:visual_assess}.

In the second analysis, we evaluate the preservation of the anatomical structure using the Struct-SSIM for each pair of original and harmonized images. Notably, we do not employ FID, LPIPS, or complete SSIM for this evaluation, as the degree of harmonization varies across the different models, as discussed in Section \ref{subsubsec:struct_metrics}. We analyze the transfer of scanner-specific characteristics using the JSD, HD, and WD metrics. These metrics are calculated based on mean voxel intensity distributions from the 10 test scanners and statistical significance is determined through paired bootstrap t-tests between pre- and post-harmonization, as outlined in Section~\ref{subsec:metrics_and_stat_tets}. We assess the performance of the proposed procedure against competing models using two distict harmonization approaches: (1) transferring images to the \textit{scanner-free} space and (2) transferring images to one of the training scanners, selecting the Gyroscan Intera scanner as the reference due to its largest representation in the training dataset. The results are detailed in Section~\ref{subsubsec:preservation_and_transfer}.

For the third analysis, we compare the proposed approach against benchmark models using the traveling subject dataset. We compute the pairwise SSIM for each subject across images acquired with different scanners, both before and after harmonization. To evaluate the statistical significance of the harmonization effect, we conduct a bootstrap t-test and report the 95\% confidence interval (95\% CI) for the difference in SSIM values between pre- and post-harmonization images, focusing solely on the proposed model’s performance when transferring images to the \textit{scanner-free} space.  The results are reported in Section~\ref{subsubsec:travel_sbj}.

\begin{table}[H]
\footnotesize
\scriptsize
\centering
\begin{tblr}{@{} >{\raggedright\arraybackslash} p{1.5cm} >{\raggedright\arraybackslash}p{3.0cm} >{\raggedright\arraybackslash}p{2.2cm} >{\raggedright\arraybackslash}p{1.8cm} >{\raggedright\arraybackslash}p{1.3cm} >{\centering\arraybackslash}p{1.2cm} @{}}
\hline[0.5pt]
\textbf{}  & \textbf{Scanner Model} & \textbf{Manufacturer} & \textbf{Dataset} & \textbf{Img. \#} & \textbf{Total} \\ 
\hline \hline
\SetCell[r=10]{c,3.0cm} \textbf{Test} & Trio & SIEMENS & SALD  & 494/494 & \SetCell[r=10]{c,1.2cm} 796  \\
&  Achieva dStream  & Philips & RIN & 17/17 \\
&  DISCOVERY MR750 & GE & RIN & 23/23  \\
&  Ingenia CX & Philips & RIN & 23/23  \\
& Prisma & SIEMENS & RIN  & 40/40  \\
&  Skyra & SIEMENS & RIN & 12/12 \\
&  Skyra Fit & SIEMENS & RIN & 3/3 \\
&  Triotim & SIEMENS & PPMI & 41/41   \\
&  Gyroscan Intera & Philips & IXI & 72/322 \\
&  Unknown & GE & IXI & 74/74 \\
\hline[0.5pt]
\end{tblr}
\caption{Description of the healthy controls test datasets, including details on scanner types, manufacturers, source datasets, and number of images.}
\label{tab:test_dataset}
\end{table}

\subsection{Downstream Analysis}
\label{subsec:down_analysis}
This section describes the downstream analysis conducted to evaluate the impact of harmonization on extracting biologically meaningful information from MRI data, comparing the proposed model with benchmark methods. We specify the datasets used for each analysis and detail the procedure. The results are presented in Section \ref{subsec:results_down_analysis}. Specifically, we perform a series of downstream analyses to assess whether harmonization enhances the reliability and predictive power of MRI-based biomarkers:
\begin{enumerate}
    \item \textbf{Age Prediction:} We explore whether harmonization enhances the accuracy of predicting personal age based on MRI-derived features.
    \item \textbf{Reduction of Inter-Scanner Variability:} We quantify how harmonization reduces variability in MRI-derived brain volumes caused by differences in scanning protocols and equipment.
    \item \textbf{AD vs. Healthy Classification:} We examine whether harmonization improves the ability to distinguish between AD patients and healthy controls, focusing on predictive performance.
    \item \textbf{Diagnosis Prediction:} We investigate the impact of harmonization on classifying mild cognitive impairment (MCI) due to AD and AD dementia.
\end{enumerate}
These analyses are designed to assess the degree to which harmonization improves the consistency and robustness of MRI-based biomarkers, ultimately enhancing their clinical and research applications. The volumetric variables used in these analyses were extracted using a \href{https://github.com/alessioc17/Pipeline-Freesurfer.git}{custom pipeline} implemented in FreeSurfer \cite{FischlSalat2002}.

\subsubsection{Age Prediction Task}
\label{subsubsec:age_pred_task}
The objective of the analysis is to determine whether harmonization enhances the accuracy of personal age prediction based on MRI-derived features. Specifically, we employed a simple linear model to predict age using a set of volumetric variables extracted from MR images. The variables selected are those most strongly associated with aging  \cite{sele2020decline, jernigan2001effects, walhovd2005effects}, including total gray matter volume (TGV), subcortical gray volume (SGV), supratentorial volume, cortex volume (CV), cerebral white matter volume, left lateral ventricle volume, left hippocampus volume (LHV), left amygdala volume, and left putamen volume (LPV). We considered the test images from healthy controls described in Table~\ref{tab:test_dataset}.
The model was applied to volumes extracted from raw MRI scans before any preprocessing or harmonization, as well as after applying each harmonization method under evaluation. To ensure robust comparisons, we perform 10-fold cross-validation and report the mean and standard deviation of the coefficient of determination (R²), the root mean square error (RMSE), and the Bayesian Information Criterion (BIC). The results are presented in Section \ref{subsubsec:results_age_pred_task}).

\subsubsection{Inter-Scanner Variability in MRI-Derived Brain Volumes}
\label{subsec:inter-scan}
We aimed to evaluate the extent to which harmonization reduces inter-scanner variability in MRI-derived brain volumes. Specifically, we focused on a subset of the variables used in Section~\ref{subsubsec:age_pred_task}, consisting of five volume variables extracted from MR images: TGV, SGV, CV, LHV, and LPV. To quantify the reduction in inter-scanner variability, we employed linear mixed-effects models (LMM) to predict these volumes based on age, treating scanner groups as random effects. 
The LMM is defined as follows:
\[
y_{ij} = \beta_0 + \beta_1 x_{ij} + u_j + \epsilon_{ij}
\]
where \( y_{ij} \) and \( x_{ij} \) represent the age and volume, respectively, for individual \( i \) scanned using scanner \( j \). Here, \( \beta_0 \) is the fixed intercept, and \( \beta_1 \) is the fixed-effect coefficient for volume. The term \( u_j \) represents the random effect for scanner group \( j \), assumed to be normally distributed with variance \( \sigma_u^2 \). \( \epsilon_{ij} \) is the residual error term, assumed to be normally distributed with variance \( \sigma_\epsilon^2 \). The random effect \( u_j \) captures between-group variance, whereas the residual error \( \epsilon_{ij} \) captures within-group variance.

The model was applied to both pre-harmonization and post-harmonization data.
In either case, we calculated three metrics: (1) the intraclass correlation coefficient (ICC) to measure the proportion of variability of the dependent variable due to scanner group differences. An ICC value close to 1 suggests that most of the variability in the outcome is due to differences between scanner groups, indicating a strong group-level influence on the dependent variable. This implies that a substantial portion of the variance in the predicted age is attributable to the scanner, reducing the reliability of volume as a biomarker of age. Conversely, an ICC value near 0 implies that variability is primarily within groups, suggesting that scanner effects have minimal impact on the outcome; (2) the marginal $R^2$ ($R^2_m$), which reflects how much of the total variance is explained by the fixed effects, excluding the contribution of the random effects; and (3) the BIC for both the LMM and the corresponding linear model (LM) that do not consider explicitly the random effects. We report the difference in BIC between the LM and the LMM (denoted as $\Delta$BIC). A lower $\Delta$BIC indicates that incorporating the scanner variable as a random effect yields minimal improvement in model fit. A higher $\Delta$BIC suggests a significant contribution of the scanner variable as a random effect. 
For this analysis, we considered the test images from healthy controls described in Table~\ref{tab:test_dataset}, excluding scanners with fewer than 15 images. Thus, $j \in$ \{Achieva dStream; DISCOVERY MR750; Ingenia CX; Prisma; Gyroscan Intera; Unknown (IOP); Triotim (PPMI); Triotim (SALD)\}.
The results are presented in Section \ref{subsec:results_inter-scan}.

\subsubsection{Classification of Alzheimer's Disease (AD) vs. Healthy Patients}
\label{subsubsec:classif_sick_vs_healthy}
We evaluated the effectiveness of harmonization in enhancing the biological information for a classification task distinguishing between healthy and AD patients. The variability introduced by the scanner effects can negatively affect classifier training, leading to a less accurate model. To evaluate this, we selected 102 MR images of healthy patients from the RIN, IXI, and PPMI test datasets (Table~\ref{tab:test_dataset}) and 102 images of AD patients from the NeuroArtP3 and ADNI3 datasets, both before and after harmonization. We trained a vanilla 3D CNN classifier 10 times with 10 different random splits, each composed of 132 MR images for training and 72 for testing. The splits were chosen to maintain an equal number of healthy control and AD images in both the training and test sets. For both our model and the two competing approaches, we employed the same classifier architecture and identical training configurations in each iteration to ensure a fair and consistent comparison. The results are presented in Section \ref{subsubsec:results_classif_sick_vs_healthy}).

\subsubsection{Diagnosis Prediction}
\label{subsubsec:diagn-pred}
We explored the effect of harmonization on classification performance in distinguishing between  MCI and AD dementia. The dataset used for this analysis includes the 41 subjects from the NeuroArtP3 dataset. We assessed performance by comparing volumes extracted from the 41 raw MRI scans before harmonization with volumes processed through preprocessing and harmonization using each of the evaluated methods. To include as much information as we could,  we considered 57 volume-related variables from the brain regions extracted from the images. To reduce dimensionality, we applied principal component analysis (PCA) and retained the number of components that explain 70\% of the variance, which provides an optimal trade-off between variance explained and model complexity across all scenarios. We then use the principal components (PCs) for logistic regression modeling of the classification task. For performance evaluation, we report the area under the ROC curve (AUC), which quantifies the model's ability to differentiate between MCI and AD dementia. An AUC of 0.5 corresponds to random performance, while an AUC of 1.0 indicates perfect classification. Higher AUC values reflect better performance in distinguishing the two conditions. The results are presented in Section \ref{subsubsec:results_diagn-pred}).

\section{Results}
\label{sec:results}

\subsection{Ablation Study Results}
\label{subsec:results_ablation_study}
We present the results of the ablation study as provided in Table~\ref{tab:ablation_study_results}. Before harmonization, the mean and standard deviation of the JSD values were $0.17 \pm 0.08$ and thus we reject the AD-test null hypothesis (p-value $\ll 0.05$). In Table~\ref{tab:ablation_study_results}, for each model setup, considering data after harmonization, we indicate whether the null hypothesis of the AD-test is accepted or rejected, with acceptance suggesting the effectiveness of harmonization. The results of the ablation study show that all tested components are crucial for enhancing the model's performance. After harmonization, the similarity between distributions across all models increases significantly in terms of JSD, resulting in comparable means and standard deviations. For all evaluated models, except the one without attention layers, we accept the null hypothesis of the AD test, indicating successful harmonization. Regarding the quality of generated images and structural preservation, the model incorporating all components performs the best, as evidenced by higher SSIM scores and lower LPIPS and FID scores. In addition, the ablation study reveals substantial improvements over the baseline DISARM in both image quality and structural preservation. The findings also confirm that the evaluation metrics effectively capture different aspects of the generated images. For instance, while the model without latent loss has a lower SSIM than the one without attention layers, it achieves a better FID score.

\begin{table*}[htbp]
\tiny
\renewcommand{\arraystretch}{1.3}
\centering
\begin{tblr}{@{} m{1.5cm} >{\centering\arraybackslash}m{0.3cm} m{0.3cm} >{\centering\arraybackslash}m{0.4cm} >{\centering\arraybackslash}m{0.4cm} >{\centering\arraybackslash}m{0.4cm} >{\centering\arraybackslash}m{1.5cm} >{\centering\arraybackslash}m{0.6cm} >{\centering\arraybackslash}m{1.2cm} >{\centering\arraybackslash}m{1.6cm} >{\centering\arraybackslash}m{1.1cm} >{\centering\arraybackslash}m{0.6cm} @{}}
\hline[0.5pt]
\textbf{Setup} & \textbf{$\boldsymbol{L^{sf}}$} & \textbf{$\boldsymbol{L^{KL}}$} & \textbf{$\boldsymbol{L^{lat}}$} & \textbf{Att.\\ Lay.} & \textbf{Part.\\ Vol.} & \textbf{SSIM} & \textbf{FID} & \textbf{LPIPS} & \textbf{JS-div (Post)} & \textbf{AD-test} & 
\\  
 \hline \hline
DISARM++    & {\ding{51}} & {\ding{51}} & {\ding{51}}& {\ding{51}} & {\ding{51}} & $0.983 \pm 0.006$ & $18.6$ & $0.10 \pm 0.02$ & $0.008 \pm 0.002$ & {\ding{51}} 
  \\
w/o $L^{\text{sf}}$    & {\ding{55}} & {\ding{51}} & {\ding{51}} & {\ding{51}} & {\ding{51}} & $0.964 \pm 0.008$ & $30.5$ & $0.15 \pm 0.03$ & $0.004 \pm 0.001$ &  {\ding{51}} 
\\
w/o $L^{\text{KL}}$  & {\ding{51}}& {\ding{55}} & {\ding{51}}  & {\ding{51}}  & {\ding{51}} & $0.960 \pm 0.007$ & $35.5$ & $0.14 \pm 0.03$ & $0.003 \pm 0.002$ & {\ding{51}} 
\\
w/o $L^{\text{lat}}$   & {\ding{51}}  & {\ding{51}} & {\ding{55}} & {\ding{51}}  & {\ding{51}}  & $0.975 \pm 0.005$ & $25.4$ & $0.14 \pm 0.03$ & $0.006 \pm 0.002$ & {\ding{51}} 
\\
w/o Att.Lay. & {\ding{51}}  & {\ding{51}}  & {\ding{51}}  & {\ding{55}} & {\ding{51}}  & $0.979 \pm 0.007$ & $35.3$ & $0.13 \pm 0.02$ & $0.012 \pm 0.005$ & {\ding{55}} 
\\
DISARM & {\ding{55}} & {\ding{51}}  & {\ding{51}}  & {\ding{55}} & {\ding{55}} & $0.958 \pm 0.009$ & $35.0$ & $0.20 \pm 0.03$ & $0.007 \pm 0.002$ & {\ding{51}} 
\\
\hline[0.5pt]
\end{tblr}
\caption{Summary of the ablation study results, with each row representing a different model configuration and its corresponding evaluation results.}
\label{tab:ablation_study_results}
\end{table*}

\subsection{Harmonization Results}
\label{subsec:results_harmonization}
This section presents the results of the harmonization assessments described in Section~\ref{subsubsec:evaluation_setup}.

\subsubsection{Visual Assessment of Harmonization}
\label{subsubsec:visual_assess}
\begin{figure}[H]
    \includegraphics[width=\linewidth]{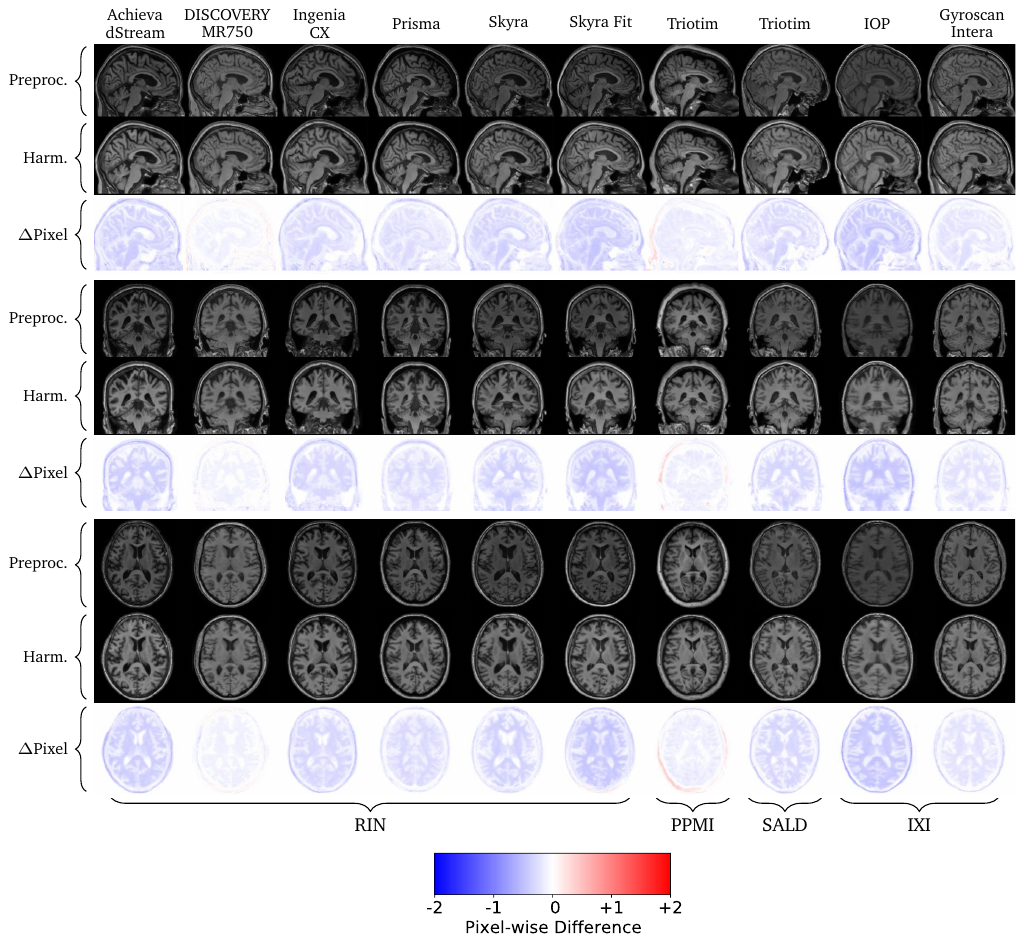}
        \caption{DISARM++ harmonization visual assessment. The figure displays slices from the axial, coronal, and sagittal dimensions for 10 original images — one per test scanner — alongside their corresponding harmonized slices. Heatmaps illustrate the pixel-wise differences between the harmonized images and their original counterparts.}
    \label{fig:disarm_harm_imgs_with_heatmaps}
\end{figure}

\begin{figure}[H]
    \includegraphics[width=\linewidth]{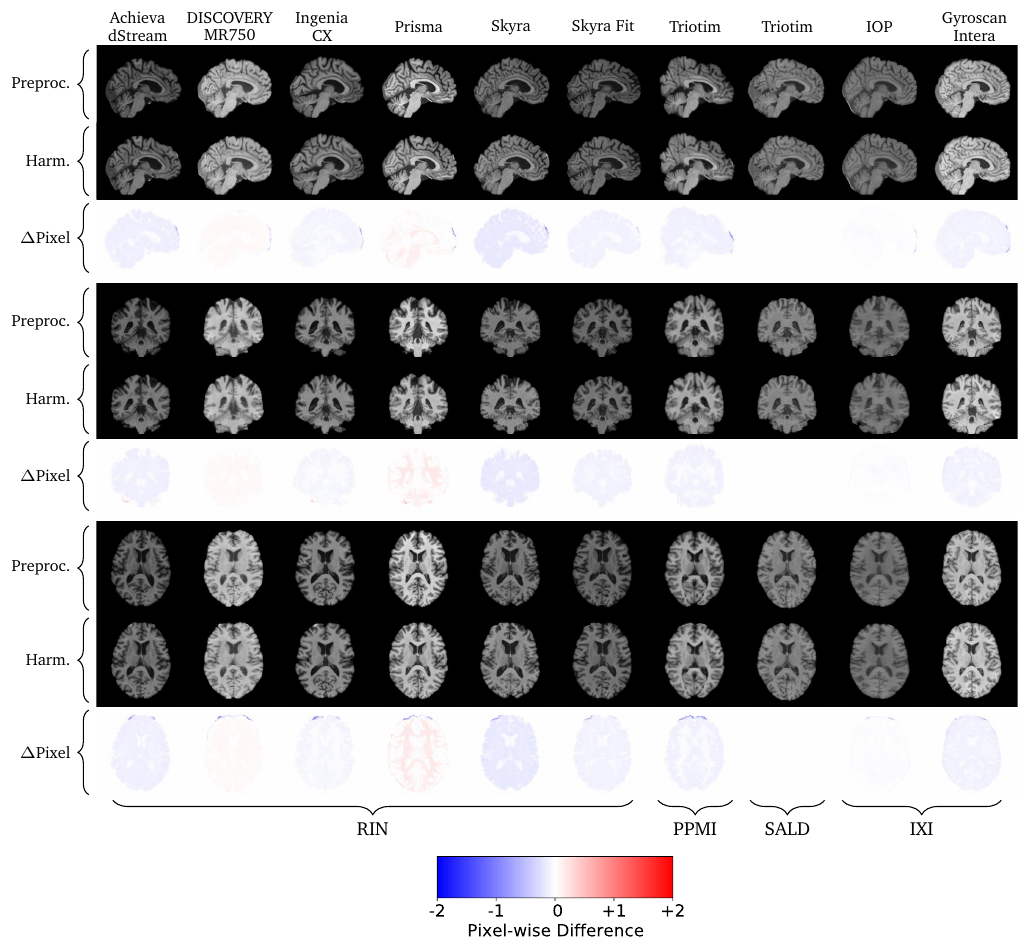}
        \caption{IGUANe harmonization visual assessment. The figure displays slices from the axial, coronal, and sagittal dimensions for 10 original images — one per test scanner — alongside their corresponding harmonized slices. Heatmaps illustrate the pixel-wise differences between the harmonized images and their original counterparts.}
    \label{fig:iguane_harm_imgs_with_heatmaps}
\end{figure}

\begin{figure}[H]
    \includegraphics[width=\linewidth]{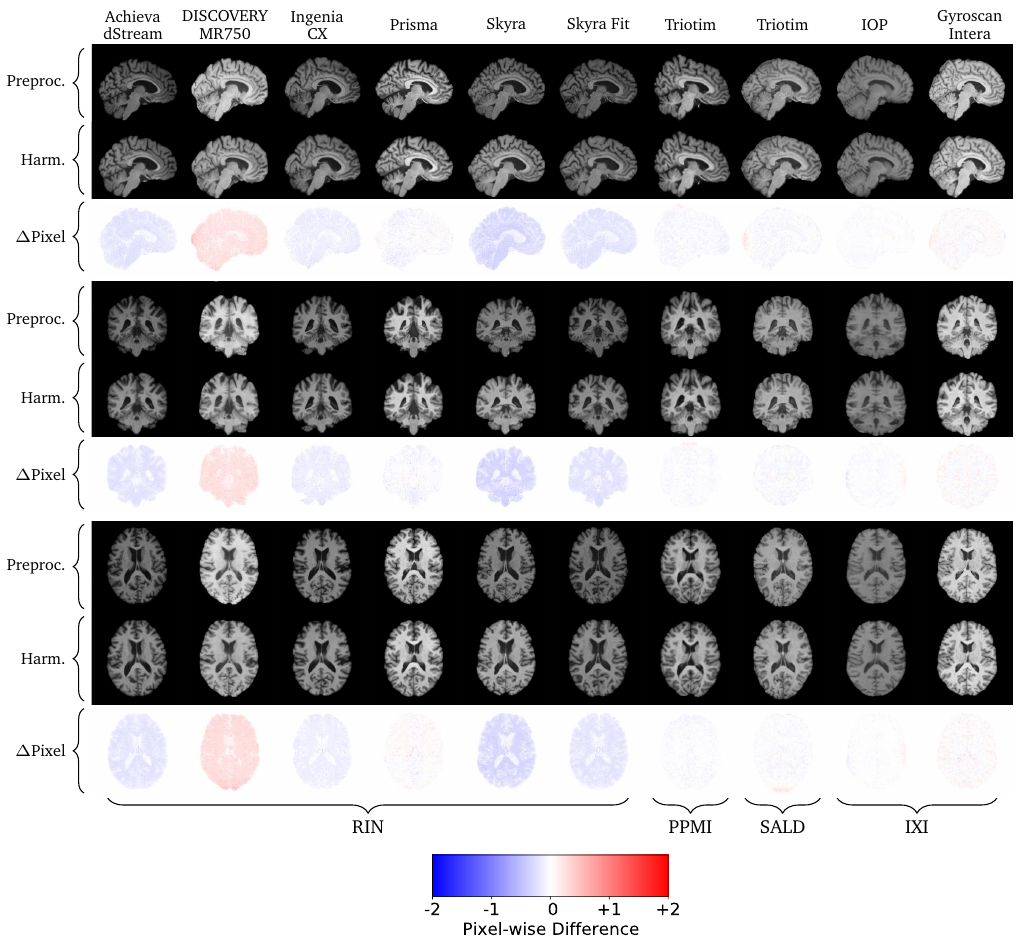} 
        \caption{STGAN harmonization visual assessment. The figure displays slices from the axial, coronal, and sagittal dimensions for 10 original images — one per test scanner — alongside their corresponding harmonized slices. Heatmaps illustrate the pixel-wise differences between the harmonized images and their original counterparts.}
    \label{fig:stgan_harm_imgs_with_heatmaps}
\end{figure}

Figures \ref{fig:disarm_harm_imgs_with_heatmaps}, \ref{fig:iguane_harm_imgs_with_heatmaps}, and \ref{fig:stgan_harm_imgs_with_heatmaps}  present the visualizations described in Section~\ref{subsubsec:evaluation_setup}. DISARM++ demonstrates a significantly stronger harmonization effect than both IGUANe and STGAN, leading to a more uniform visual appearance across all directions in the images acquired with the ten different scanners. This is further emphasized by the heatmaps, which reveal a stronger effect for DISARM++, followed by STGAN, while IGUANe's heatmaps are the least pronounced.

\subsubsection{Assessment of Anatomical Structure Preservation and Scanner Characteristics Transfer}
\label{subsubsec:preservation_and_transfer}
In this section, we evaluate the proposed model's ability to preserve anatomical structures and transfer scanner-specific characteristics as outlined in Section~\ref{subsubsec:evaluation_setup}, comparing it with the benchmark models.

Regarding the preservation of anatomical structure, Table~\ref{tab:comparison_eval} presents the means and standard deviations of Struct-SSIM across all test images. Both IGUANe and STGAN achieve slightly higher scores than DISARM++. However, when assessing anatomical preservation, the presence of large black borders in skull-stripped images—compared to non-skull-stripped images—presents a notable challenge. These black regions, resulting from the removal of non-brain tissues during skull-stripping, are uniform and low-intensity, which can inflate similarity scores. Consequently, the Struct-SSIM may yield high values as it considers the black borders structurally similar between the original and harmonized images. This likely explains why IGUANe and STGAN exhibit slightly higher Struct-SSIM scores compared to DISARM++, even though the results remain visually similar, as shown in the previous section. 
Moreover, we deliberately avoid skull-stripping in our method for several reasons. First, since skull-stripping would need to be performed both before and after harmonization—unlike in other models—BET might inconsistently remove some non-brain areas at different stages, leading to discrepancies that negatively impact the Struct-SSIM scores. Second, skull-stripping is computationally expensive; for instance, HD-BET \cite{isensee2019automated} takes approximately ten minutes per image on a CPU, significantly increasing processing time. Finally, we choose to retain the skull to maintain the integrity of the original anatomical structures, which is particularly beneficial for applications beyond brain tissue analysis, such as studies involving head trauma or cranial deformities.

In terms of transferring scanner-specific characteristics, Table~\ref{tab:comparison_eval} summarizes the means and standard deviations of all pairwise distribution comparisons after harmonization in terms of JSD, HD, and WD. Comprehensive heatmaps displaying all pairwise values for each model before and after harmonization across all three metrics can be found in Appendix D of the Supplementary Materials. DISARM++ — including both the \textit{scanner-free} and reference scanner harmonization approaches — demonstrates the best overall performance, yielding the lowest metric values. For both harmonization strategies, the bootstrap t-test results in a p-value \( \ll 0.05 \), confirming a significant increase in distribution similarity across all three metrics. The negative confidence intervals (Table~\ref{tab:comparison_eval}) highlight the substantial increase in similarity. IGUANe shows higher means and standard deviations for JSD, HD, and WD compared to DISARM++, with bootstrap t-tests indicating a statistically significant but minor improvement in distribution similarity. Similarly, STGAN exhibits higher means and standard deviations for JSD, HD, and WD than DISARM++ but demonstrates statistically significant improvements across all three metrics — greater than IGUANe, yet still lower than DISARM++.

\begin{figure}[H]
        \centering
        \includegraphics[width=\linewidth]{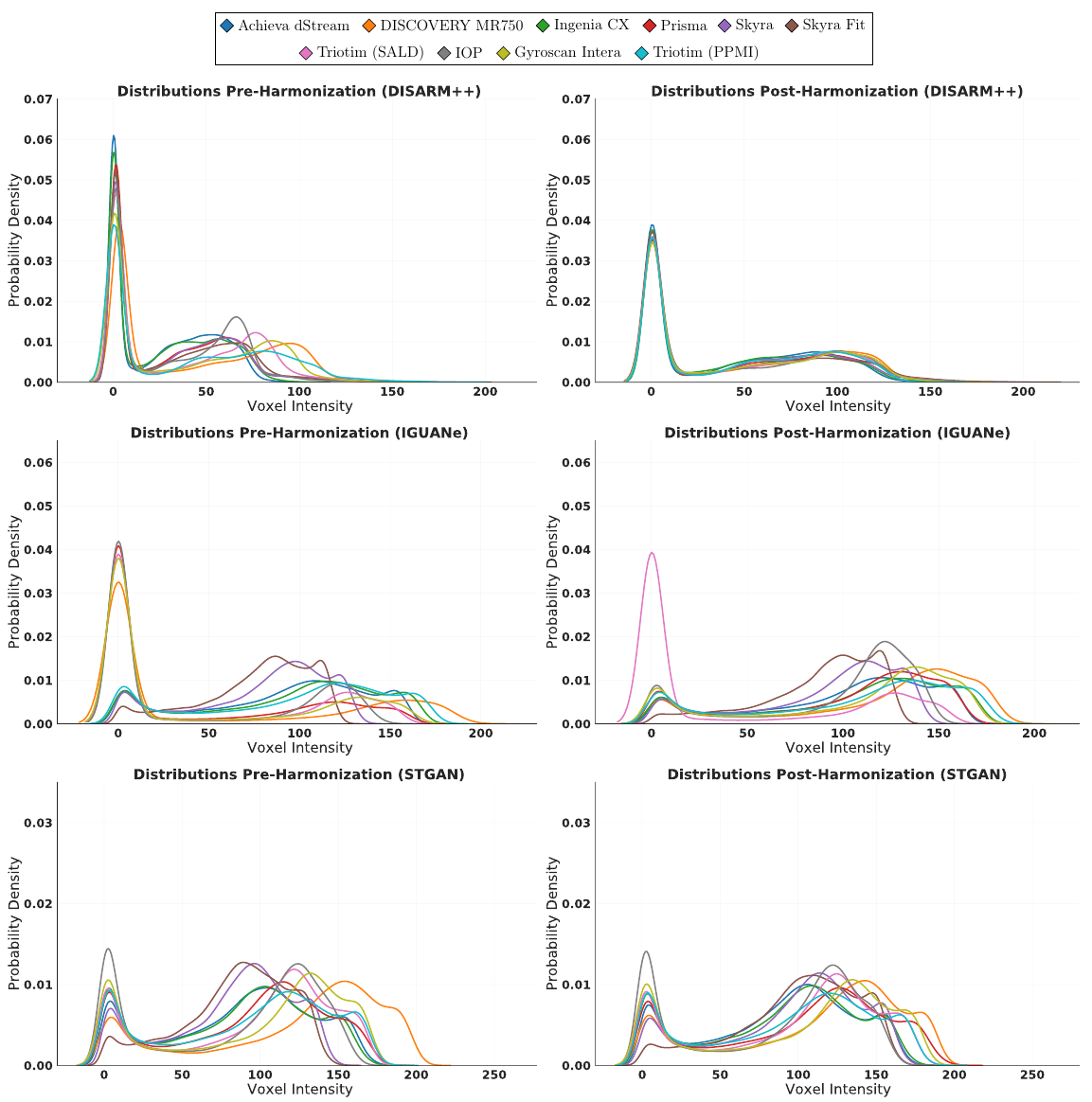}
        \caption{Comparison of mean voxel intensity distributions from the 10 test scanners before and after harmonization using DISARM++, IGUANe, and STGAN.}
        \label{fig:test_gd}
\end{figure}

Figure~\ref{fig:test_gd} visually compares the mean voxel intensity distributions for each test scanner before and after harmonization for the three models. This visual comparison supports the previously described metric results and statistical test outcomes, demonstrating that DISARM++ has a more substantial effect between pre- and post-harmonization and leads to better alignment of the distributions after harmonization.

The metric results and the visualizations of the voxel intensity distributions presented in this section further support the visual assessment discussed in the previous section.

\begin{table}[H]
\tiny
\renewcommand{\arraystretch}{1.5}
\centering
\begin{tblr}{@{} m{1.7cm} >{\centering\arraybackslash}m{2.0cm} >{\centering\arraybackslash}m{2.1cm} >{\centering\arraybackslash}m{2.0cm} >{\centering\arraybackslash}m{2.1cm} >{\centering\arraybackslash}m{2.0cm} @{}}
\hline[0.5pt]
\textbf{Metric} & \textbf{DISARM++} (\textit{Scanner-free}) & \textbf{DISARM++} (Gyroscan Intera) & \textbf{IGUANE} (SALD) & \textbf{STGAN} (Gyroscan Intera) 
\\  
\hline \hline
\textbf{Struct-SSIM} & $0.986 \pm 0.005$ & $0.986 \pm 0.005$ & $0.996 \pm 0.001$ & $0.997 \pm 0.001$ \\
\textbf{JSD} (Post) & $0.009 \pm 0.004$ & $0.010 \pm 0.004$  & $0.1281 \pm 0.1004$ & $0.0612 \pm 0.0373$ \\
\textbf{JSD} (95\% CI) & $[-0.23, -0.16]^{*}$ & $[-0.22, -0.16]^{*}$  & $[-0.04, -0.01]^{*}$ & $[-0.06, -0.02]^*$ \\ [1mm] 

\textbf{HD} (Post) & $0.154 \pm 0.038$ & $ 0.162 \pm 0.037$ & $0.6082 \pm 0.2643$ & $0.4154 \pm 0.1496$  \\
\textbf{HD} (95\% CI) & $[-0.68, -0.52]^{*}$ & $[-0.65, -0.51]^{*}$ & $[-0.09, -0.02]^{*}$ & $[-0.16, -0.05]^*$  \\ [1mm]

\textbf{WD} (Post) & $2.173 \pm 0.962$ & $ 2.059 \pm 0.880 $ & $14.329 \pm 7.891$ & $13.548 \pm 5.814$  \\
\textbf{WD} (95\% CI) & $[-6.71, -4.55]^{*}$ & $[-6.66, -4.61]^{*}$ & $[-4.05, -1.26]^{*}$ & $[-5.38, -1.71]^*$  \\ [1mm]
\hline[0.5pt]
\end{tblr}
\caption{Comparison of harmonization methods on healthy controls MR scans concerning the preservation of anatomical structures and the transfer of scanner characteristics. The asterisks denote significant values.}
\label{tab:comparison_eval}
\end{table}

\subsubsection{Traveling Subjects}
\label{subsubsec:travel_sbj}
With reference to Table \ref{tab:trav_subj}, we evaluate the proposed model and compare it to the benchmark models using the traveling subject dataset as described in Section~\ref{subsubsec:evaluation_setup}. The results show that DISARM++ leads to a significant improvement in SSIM after harmonization for all subjects in the dataset, with the 95\% CI consistently ranging from 0.13 to 0.24, indicating a robust and significant improvement. In contrast, IGUANe shows a significant increase in SSIM in only three of the seven subjects and the observed gains are considerably smaller than those achieved by DISARM++. For two subjects, there is a slight but significant decrease in SSIM post-harmonization, while for the remaining two subjects, no significant increase is observed. For STGAN, a significant improvement in SSIM is observed in all seven subjects, with 95\% CI consistently ranging from 0.02 to 0.04. These confidence intervals contain substantially smaller values compared to those of DISARM++, indicating a much less pronounced improvement. Nonetheless, STGAN consistently outperforms IGUANe, where the SSIM increase is limited to just three subjects, and the 95\% CI consistently ranges from 0.002 to 0.011.

\begin{figure}[H]
    \includegraphics[width=\linewidth]{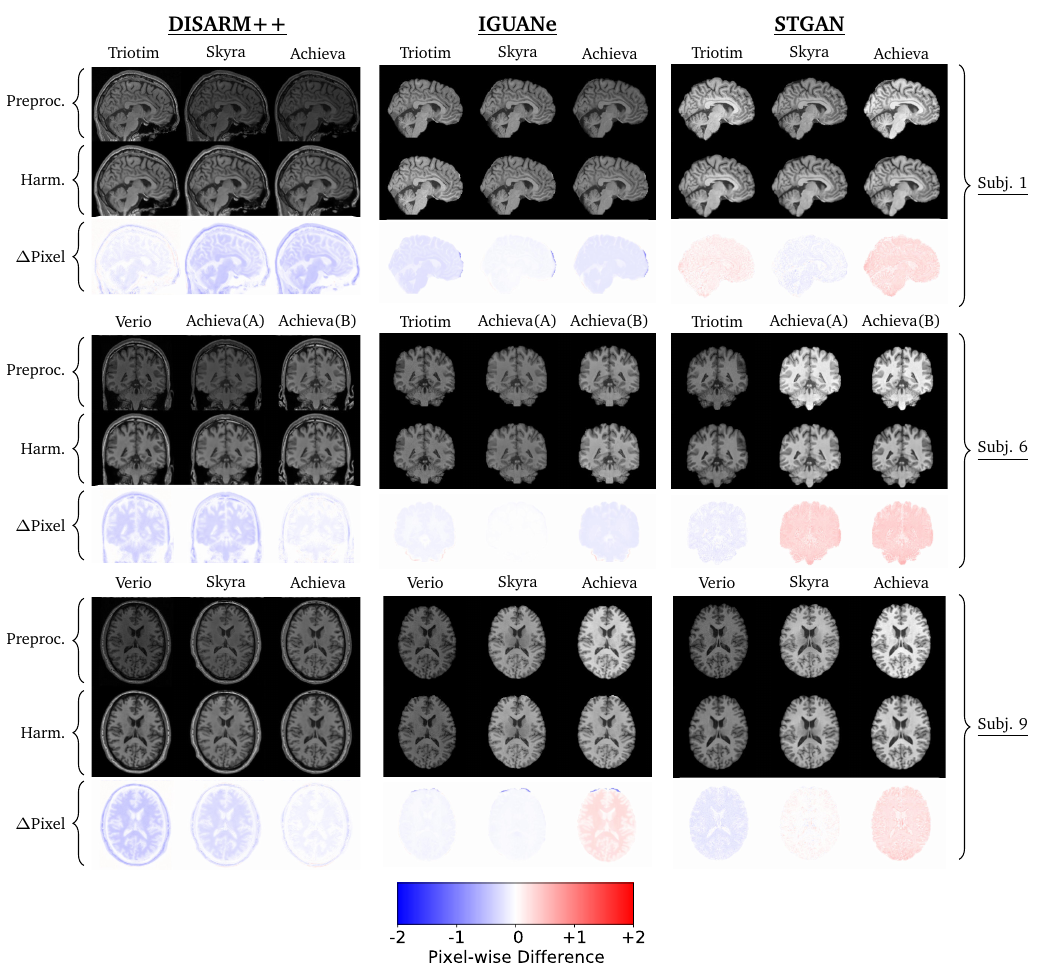} 
        \caption{Visual harmonization comparison for traveling subjects. For Subject 6, we present sagittal slices of MR images acquired with three different scanners, both before and after harmonization, along with heatmaps showing pixel-wise differences. Similarly, for Subject 1, we provide coronal slices, and for Subject 9, we display axial slices.}
    \label{fig:travel_sub_montage}
\end{figure}

Moreover, we provide a visual comparison of the harmonization results in Figure~\ref{fig:travel_sub_montage}. For subject 1, we display a sagittal slice from three MR images acquired using different scanners, showing both pre- and post-harmonization slices, along with heatmaps of pixel-wise differences. For subject 6, we present coronal slices, and for subject 9, we show axial slices. DISARM++ demonstrates a significant improvement between pre- and post-harmonization slices, resulting in harmonized slices with a much more similar visual appearance with respect to IGUANe and STGAN. This visualization further confirm the SSIM metric results.

\begin{table}[H]
\centering
\begin{minipage}{0.46\textwidth}
\tiny
\renewcommand{\arraystretch}{1.5}
\centering
\hspace{0.6cm} \textbf{\scriptsize \underline{DISARM++}} \\[0.5em]
\begin{tblr}{@{} >{\centering\arraybackslash}m{0.4cm} >{\centering\arraybackslash}m{1.5cm} >{\centering\arraybackslash}m{1.5cm} >{\centering\arraybackslash}m{2.0cm} >{\centering\arraybackslash}m{2.0cm} @{} }
\hline[0.5pt]
\textbf{Subj.} & \textbf{SSIM} (pre-harm) & \textbf{SSIM} (post-harm) & \textbf{SSIM} (95\% CI) \\  
\hline \hline
\textbf{1} & $0.633 \pm 0.119$ & $0.821 \pm 0.05$ & $[+0.160, +0.218]^{*}$ \\
\textbf{2} & $0.659 \pm 0.123$ & $0.817 \pm 0.06$ & $[+0.130, +0.186]^{*}$ \\
\textbf{3} & $0.642 \pm 0.114$ & $0.806 \pm 0.05$ & $[+0.139, +0.191]^{*}$ \\
\textbf{6} & $0.602 \pm 0.122$ & $0.785 \pm 0.06$ & $[+0.150, +0.216]^{*}$ \\
\textbf{7} & $0.605 \pm 0.138$ & $0.804 \pm 0.05$ & $[+0.155, +0.242]^{*}$ \\
\textbf{8} & $0.609 \pm 0.128$ & $0.800 \pm 0.06$ & $[+0.159, +0.219]^{*}$ \\
\textbf{9} & $0.617 \pm 0.114$ & $0.801 \pm 0.05$ & $[+0.155, +0.213]^{*}$ \\
\hline[0.5pt]
\end{tblr}
\end{minipage}
\hfill \hfill
\begin{minipage}{0.46\textwidth}
\tiny
\renewcommand{\arraystretch}{1.5}
\centering
\vspace{0.05cm}
\hspace{0.6cm} \textbf{\scriptsize \underline{IGUANe}} \\[0.5em]
\begin{tblr}{@{} >{\centering\arraybackslash}m{0.4cm} >{\centering\arraybackslash}m{1.5cm} >{\centering\arraybackslash}m{1.5cm} >{\centering\arraybackslash}m{2.0cm} >{\centering\arraybackslash}m{2.0cm} @{} }
\hline[0.5pt]
\textbf{Subj.} & \textbf{SSIM} (pre-harm) & \textbf{SSIM} (post-harm) & \textbf{SSIM} (95\% CI) \\  
\hline \hline
\textbf{1} & $0.910 \pm 0.03$ & $0.911 \pm 0.02$ & $[-0.002, +0.005]$ \\
\textbf{2} & $0.916 \pm 0.03$ & $0.922 \pm 0.02$ & $[+0.003, +0.0009]^{*}$ \\
\textbf{3} & $0.917 \pm 0.03$ & $0.917 \pm 0.03$ & $[-0.001, +0.002]$ \\
\textbf{6} & $0.922 \pm 0.02$ & $0.919 \pm 0.02$ & $[-0.005, -0.002]^{*}$ \\
\textbf{7} & $0.916 \pm 0.03$ & $0.911 \pm 0.03$ & $[-0.008, -0.002]^{*}$ \\
\textbf{8} & $0.916 \pm 0.03$ & $0.922 \pm 0.02$ & $[+0.002, +0.008]^{*}$ \\
\textbf{9} & $0.913 \pm 0.03$ & $0.921 \pm 0.02$ & $[+0.005, +0.011]^{*}$ \\
\hline[0.5pt]
\end{tblr}
\end{minipage}
\vspace{2mm} \\
\begin{minipage}{0.46\textwidth}
\tiny
\renewcommand{\arraystretch}{1.5}
\centering
\hspace{0.6cm} \textbf{\scriptsize \underline{STGAN}} \\[0.5em]
\begin{tblr}{@{} >{\centering\arraybackslash}m{0.4cm} >{\centering\arraybackslash}m{1.5cm} >{\centering\arraybackslash}m{1.5cm} >{\centering\arraybackslash}m{2.0cm} >{\centering\arraybackslash}m{2.0cm} @{} }
\hline[0.5pt]
\textbf{Subj.} & \textbf{SSIM} (pre-harm) & \textbf{SSIM} (post-harm) & \textbf{SSIM} (95\% CI) \\  
\hline \hline
\textbf{1} & $0.881 \pm 0.03$ & $0.916 \pm 0.03$ & $[+0.030, +0.040]^{*}$ \\
\textbf{2} & $0.898 \pm 0.03$ & $0.932 \pm 0.02$ & $[+0.028, +0.040]^{*}$ \\
\textbf{3} & $0.896 \pm 0.03$ & $0.919 \pm 0.03$ & $[+0.021, +0.026]^{*}$ \\
\textbf{6} & $0.886 \pm 0.04$ & $0.919 \pm 0.02$ & $[+0.028, +0.039]^{*}$ \\
\textbf{7} & $0.886 \pm 0.03$ & $0.916 \pm 0.03$ & $[+0.024, +0.035]^{*}$ \\
\textbf{8 } & $0.900 \pm 0.03$ & $0.925 \pm 0.02$ & $[+0.021, +0.029]^{*}$ \\
\textbf{9} & $0.887 \pm 0.04$ & $0.917 \pm 0.03$ & $[+0.026, +0.036]^{*}$ \\
\hline[0.5pt]
\end{tblr}
\end{minipage}
\caption{Mean SSIM values and standard deviations across all image pairs for each subject evaluated, both before and after harmonization, comparing DISARM++, IGUANe, and STGAN. The 95\% confidence intervals indicate the differences in SSIM between pre- and post-harmonization.}
\label{tab:trav_subj}
\end{table}

\subsection{Downstream Analysis Results}
\label{subsec:results_down_analysis}
This section presents the results of the downstream analysis described in Section~\ref{subsec:down_analysis}.

\subsubsection{Age Prediction Task}
\label{subsubsec:results_age_pred_task}
We present the results of the downstream analysis described in Section \ref{subsubsec:age_pred_task}, i.e., the age prediction task. With reference to Table~\ref{tab:age_pred_task}, the DISARM++ model outperforms all others, achieving the highest \(R^2\) score (\( \simeq 0.60 \, \pm \, 0.05\)), the lowest RMSE (\( \simeq 0.168 \, \pm \, 0.008 \)), and the best BIC score (\( \simeq -455.4 \, \pm \,  7.6\)). These results demonstrate DISARM++'s superior predictive accuracy and model efficiency. In contrast, the baseline raw images exhibit significantly lower performance, while STGAN provides only marginal improvements over the baseline and remains inferior to DISARM++. Finally, IGUANe performs slightly worse than the baseline.

\begin{table}[H]
\tiny
\renewcommand{\arraystretch}{1.9}
\centering
\begin{tblr}{@{} m{1.0cm} >{\centering\arraybackslash}m{2.0cm} >{\centering\arraybackslash}m{2.0cm} >{\centering\arraybackslash}m{2.0cm} >{\centering\arraybackslash}m{2.2cm} 
@{}}
\hline[0.5pt]
\textbf{Metric} & \textbf{Raw Images} & \textbf{DISARM++} (\textit{Scanner-free}) & \textbf{IGUANE} (SALD) & \textbf{STGAN} (Gyroscan Intera) 
\\  
\hline \hline
$\boldsymbol{R^2}$ & $0.5209 \pm 0.1126$  & $\boldsymbol{0.6042 \pm 0.0495}$ & $0.4879 \pm 0.0665$ & $0.5290 \pm 0.1177$ &  \\
\textbf{RMSE} & $0.1873 \pm 0.0259$ & $\boldsymbol{0.1684 \pm 0.008}$ & $0.1941 \pm 0.018$ & $0.1871 \pm 0.034$  &  \\
\textbf{BIC} & $-353.6 \pm 12.6$ & $\boldsymbol{-455.4 \pm 7.6}$ & $-266.6 \pm 12.9$ & $-374.3 \pm 10.0$  &  \\
\hline[0.5pt]
\end{tblr}
\caption{Comparison of models on age prediction task. Bold font denotes the best results.}
\label{tab:age_pred_task}
\end{table}

\subsubsection{Inter-Scanner Variability in MRI-Derived Brain Volumes}
\label{subsec:results_inter-scan}

 \begin{table}[H]
\tiny
\renewcommand{\arraystretch}{1.5}
\centering
\begin{tblr}{@{} >{\raggedright\arraybackslash}m{0.7cm} >{\raggedright\arraybackslash}m{1.1cm} >{\centering\arraybackslash}m{1.8cm} >{\centering\arraybackslash}m{2.2cm} >{\centering\arraybackslash}m{1.8cm} >{\centering\arraybackslash}m{2.2cm} 
@{}}
\hline[0.5pt]
\textbf{Var.} & \textbf{Metrics} & \textbf{Raw Images} & \textbf{DISARM++} (\textit{Scanner-free}) & \textbf{IGUANe} (SALD) & \textbf{STGAN} (Gyroscan Intera) 
\\
\hline \hline
 \SetCell[r=3]{c,3.5cm} \textbf{TGV} & \textbf{ICC} (\%) & $60.19$ & $\boldsymbol{11.43}$ & $55.91$ & $27.59$ & \\
 & $\boldsymbol{R_m}$  (\%) & $22.37$ & $\boldsymbol{50.36}$ & $32.65$ & $41.81$   \\
& \textbf{$\Delta$BIC}  &  $+ \, 287$   & $\boldsymbol{+ \, 18}$ & $+ \, 417$ & $+ \, 95$ \\
\hline
 \SetCell[r=3]{c,3.5cm} \textbf{SGV} & \textbf{ICC} (\%) & $54.28$ & $\boldsymbol{7.21}$ & $58.60$ & $15.39$ \\
 & $\boldsymbol{R_m}$ (\%) & $19.67$ & $29.17$ & $\boldsymbol{32.62}$ & $26.65$  \\
& \textbf{$\Delta$BIC}  &  $+ \, 246$   & $\boldsymbol{- \, 5}$ & $+ \, 383$ & $+ \, 25$ \\
\hline
 \SetCell[r=3]{c,3.5cm} \textbf{CV} & \textbf{ICC} (\%) & $57.84$ & $\boldsymbol{16.46}$ & $49.21$ & $26.88$ \\
 & $\boldsymbol{R_m}$ (\%) & $22.87$ & $\boldsymbol{51.12}$ & $27.81$ & $40.28$   \\
& \textbf{$\Delta$BIC}  &  $+ \, 271$   & $\boldsymbol{+ \, 34}$ & $+ \, 382$ & $+ \, 93$ \\
\hline
 \SetCell[r=3]{c,3.5cm} \textbf{LHV} & \textbf{ICC} (\%) & $42.61$ & $\boldsymbol{12.80}$ & $40.52$ & $24.89$ \\
 & $\boldsymbol{R_m}$ (\%) & $11.64$ & $\boldsymbol{13.44}$ & $12.91$ & $3.71$   \\
& \textbf{$\Delta$BIC}  &  $+ \, 204$   & $\boldsymbol{+ \, 22}$ & $+ \, 223$ & $+ \, 113$ \\
\hline
 \SetCell[r=3]{c,3.5cm} \textbf{LPV} & \textbf{ICC} (\%) & $47.28$ & $\boldsymbol{13.08}$ & $39.13$ & $23.39$ \\
 & $\boldsymbol{R_m}$ (\%) & $16.30$ & $\boldsymbol{25.99}$ & $12.85$ & $15.79$   \\
& \textbf{$\Delta$BIC}  &  $+ \, 203$   & $\boldsymbol{+ \, 17}$ & $+ \, 262$ & $+ \, 66.39$ \\
\hline[0.5pt]
\end{tblr}
\caption{Comparison of inter-scanner variability in volumes extracted from images before preprocessing and harmonization, and after applying all harmonization models. The $\Delta$BIC represents the difference between the metric computed without and with random effects. Bold font denotes the best results.}
\label{tab:inter_scan_var}
\end{table}

\begin{figure}[H]
    \includegraphics[width=\linewidth]{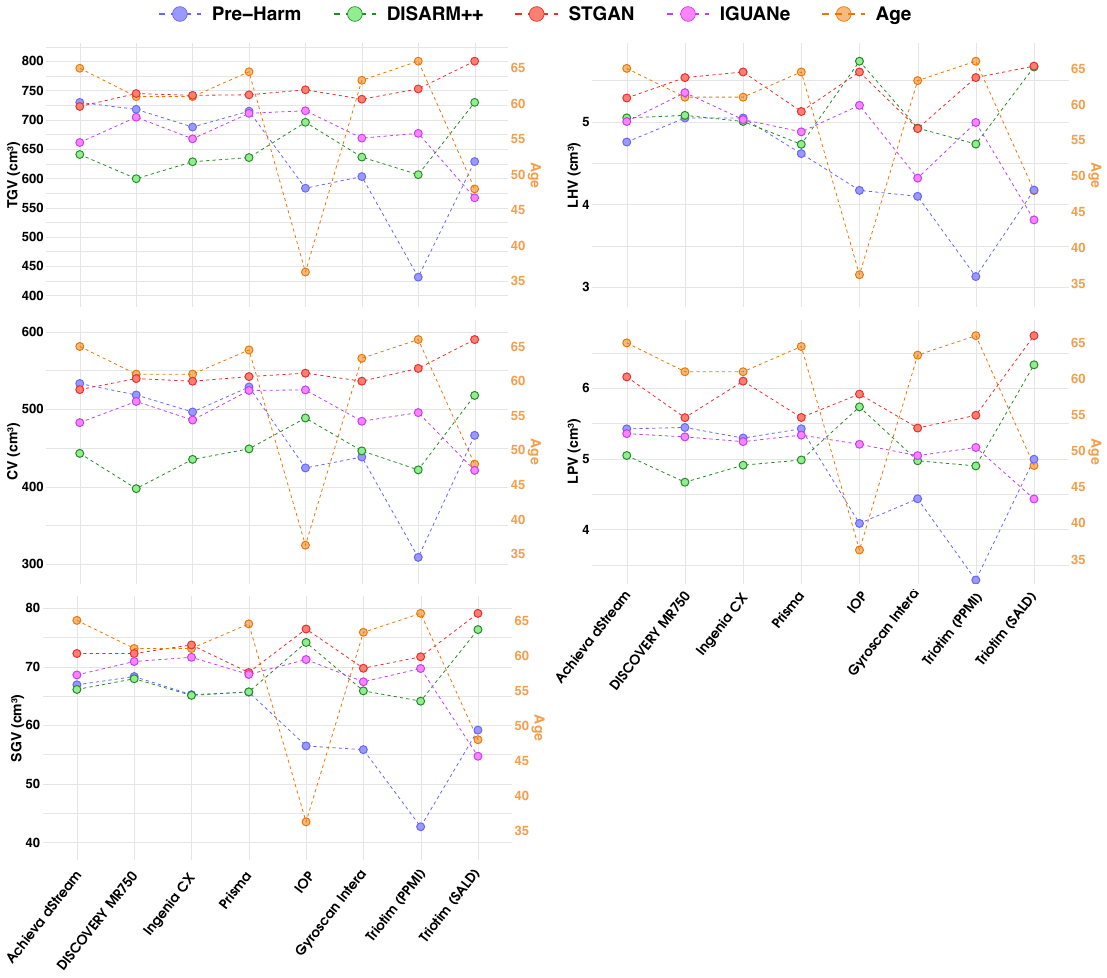} 
        \caption{For each volume variable considered, we present the median volumes for each scanner before preprocessing or harmonization (orange), along with those obtained from the three competing models: DISARM++ (green), STGAN (red), and IGUANe (purple). In addition, we report the median age for each test scanner (blue).}
    \label{fig:inter_scan}
\end{figure}

In this section, we present the results of the downstream analysis described in Section \ref{subsec:inter-scan}, concerning the inter-scanner variability assessment in MRI-derived brain volumes. The three metrics for all five variables are presented in Table~\ref{tab:inter_scan_var}. DISARM++ harmonization, across all five variables, significantly reduces inter-scanner variability, as evidenced by a substantial decrease in the ICC (e.g., ICC for TGV drops from $\simeq 60\%$ to $\simeq 11\%$). Furthermore, harmonization increases the proportion of variance explained by fixed effects alone (e.g., $R_m$ for TGV rises from $\simeq 22\%$ to $\simeq 50\%$). Before preprocessing, the BIC for the models with random effects are significantly lower than that for the models without, indicating that random effects are essential for a better model fit (e.g., $\Delta$BIC $\simeq +287$ for TGV). Following DISARM++ harmonization, the BIC for the models without random effects become comparable to, or even lower than, that for the models with random effects, suggesting that random effects are no longer crucial for predicting the outcome (e.g., $\Delta$BIC $\simeq -5$ for SGV). In contrast, both STGAN and IGUANe harmonization yield a smaller reduction in inter-scanner variability, explain a lower proportion of variance via fixed effects (except for SGV in the case of IGUANe), and consistently result in worse $\Delta$BIC values across all five variables compared to DISARM++. For instance, the ICC for TGV is approximately 28\% under STGAN harmonization and 56\% under IGUANe, compared to only 11\% with DISARM++; the $R_m$ is around 42\% with STGAN and around 33\% with IGUANe, versus 50\% with DISARM++; and the $\Delta$BIC is roughly +95 for STGAN and +419 for IGUANe, compared to +18 for DISARM++. Moreover, Figure \ref{fig:inter_scan} illustrates the median volumes for each scanner across the analyzed volume variables, both before harmonization and after applying the three competing models. Additionally, the figures depict the median age for each test scanner, as brain region volumes generally decline with aging. Among the scanners, IOP and Triotim (SALD) include the youngest individuals (IQR = [30, 50] and IQR = [25, 59], respectively), while Triotim (PPMI) has the oldest (IQR = [64, 71]). In the TGV plot, raw image volumes (blue) reveal that the IOP and Triotim (SALD) scanners have lower median volumes (IQR = [547, 654] and IQR = [588, 678], respectively) compared to Achieva dStream (IQR = [668, 763]), DISCOVERY MR750 (IQR = [691, 750]), Ingenia CX (IQR = [662, 743]), and Prisma (IQR = [674, 758]), which correspond to scanners with higher median ages. This observation is counterintuitive, as total brain volume (TGV) typically decreases with age. After applying DISARM++ (green), the IOP and Triotim (SALD) scanners exhibit the highest median volumes (IQR = [654, 724] and IQR = [695, 766], respectively), aligning with the expected trend of age-related volume reduction. The median values for the other scanners remain comparable across groups, reflecting their similar age distributions. Conversely, STGAN (red) produces uniform median values across all scanners, indicating that the model homogenizes volumes regardless of age distribution differences. For instance, IOP (IQR = [719, 773]) and Triotim (PPMI) (IQR = [733, 774]) display nearly identical median volumes despite their age disparities. Similarly, IGUANe (purple) results in comparable median volumes for IOP (IQR = [692, 737]), Ingenia CX (IQR = [651, 683]), and DISCOVERY MR750 (IQR = [679, 721]), despite differences in median age. Moreover, IGUANe yields the lowest median volume values for Triotim (SALD), even though this cohort has the second youngest median age among all scanners, following IOP.

\subsubsection{Classification of Alzheimer's Disease (AD) vs. Healthy Patients}
\label{subsubsec:results_classif_sick_vs_healthy}
\begin{figure}[H]
    \includegraphics[width=\linewidth]{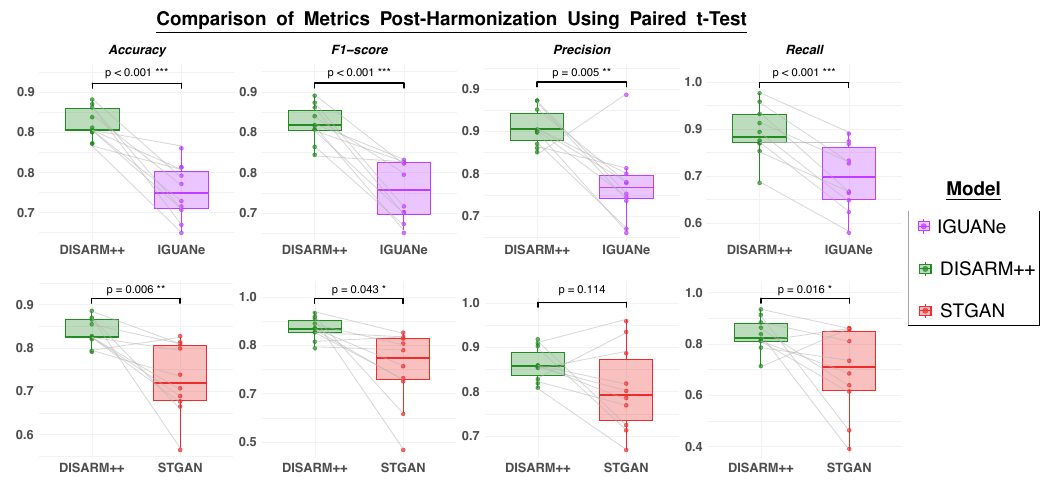} 
        \caption{Comparison of metrics between DISARM++ and IGUANe and between DISARM++ and STGAN, using a paired bootstrap t-test, post-harmonization harmonization across the 10 random splits.}
    \label{fig:fig_class}
\end{figure}

In this section, we present the results of the downstream analysis described in Section \ref{subsubsec:classif_sick_vs_healthy} investigating the discrimination power between pathological and healthy subjects. Table~\ref{tab:table_class} summarizes the classifier's performance in terms of accuracy, precision, recall, and F1-score. DISARM++ harmonization outperforms the other methods across all four metrics. Additionally, Figure~\ref{fig:fig_class} displays boxplots showing the distributions of these metrics after harmonization for each evaluated model. Paired bootstrap t-tests show that DISARM++ significantly outperforms IGUANe across all four metrics. Compared to STGAN, DISARM++ shows significantly better precision, F1 score, and recall, although no significant difference in precision is observed.

\begin{table}[H]
\tiny
\renewcommand{\arraystretch}{1.5}
\centering
\begin{tblr}{@{} >{\raggedright\arraybackslash}p{1.2cm} >{\centering\arraybackslash}p{1.4cm} >{\centering\arraybackslash}p{1.4cm} >{\centering\arraybackslash}p{1.4cm} >{\centering\arraybackslash}p{1.4cm} >{\centering\arraybackslash}p{1.4cm} >{\centering\arraybackslash}p{1.4cm}  >{\centering\arraybackslash}p{1.4cm} >{\centering\arraybackslash}p{1.4cm}@{}}
\cline{1-7}
 & \SetCell[c=2]{c, 1.6cm}{\textbf{DISARM++} \\ (\textit{Scanner-free})} & & \SetCell[c=2]{c, 1.6cm}{\textbf{IGUANe} \\ (SALD)} & & \SetCell[c=2]{c, 2.2cm}{\textbf{STGAN} \\ (Gyroscan Intera)} & 
 \\
 & \textbf{Pre} & \textbf{Post} & \textbf{Pre} & \textbf{Post} & \textbf{Pre} & \textbf{Post} 
 \\
\hline \hline
\textbf{Accuracy} & $0.807 \pm 0.05$ & $0.858 \pm 0.03$ & $0.731 \pm 0.05$ & $0.738 \pm 0.05$ & $0.751 \pm 0.06$ & $0.757 \pm 0.07$  \\
\textbf{Precision} & $0.784 \pm 0.07$ & $0.859 \pm 0.03$ & $0.740 \pm 0.08$ & $0.744 \pm 0.07$ & $0.742 \pm 0.06$ & $0.804 \pm 0.09$   \\
\textbf{Recall} & $0.861 \pm 0.09$ & $0.861 \pm 0.07$ & $0.733 \pm 0.101$ & $0.742 \pm 0.09$ & $0.775 \pm 0.114$ & $0.700 \pm 0.19$  \\
\textbf{F1-score} & $0.816 \pm 0.05$ & $0.858 \pm 0.03$ & $0.730 \pm 0.06$ & $0.737 \pm 0.05$ & $0.754 \pm 0.07$ & $0.731 \pm 0.11$   \\
\hline [0.5pt]
\end{tblr}
\caption{Comparison of the classifier quality from all models before and after harmonization.}
\label{tab:table_class}
\end{table}

\subsubsection{Diagnosis Prediction}
\label{subsubsec:results_diagn-pred}
This section presents the results of the downstream analysis described in Section \ref{subsubsec:diagn-pred} aiming at the prediction of diagnosis, i.e., AD or MCI. The LR model trained on raw MRI volumes (without preprocessing or harmonization) achieves an AUC of 0.6284. After applying DISARM++ harmonization, the AUC significantly improves to 0.9459, highlighting a substantial enhancement in classification performance. In comparison, STGAN harmonization increases the AUC to 0.7353, reflecting a more moderate improvement. IGUANe achieves an AUC of 0.8176, outperforming STGAN but not reaching the performance of DISARM++.

\section{Discussion}
\label{sec:discussion}
In this study, we introduced a novel approach to harmonizing T1-weighted MR images acquired from different scanners. Unlike existing harmonization techniques that focus on standardizing image-derived features, our method directly harmonizes the brain MRI. This ensures that downstream features extracted from the harmonized images are inherently consistent, enhancing their reliability for various analysis. Our image-based approach allows for image transfer from different scanners in two distinct ways: (1) transferring images to a \textit{scanner-free} space, ensuring consistent appearances regardless of the original scanner source; (2) mapping images to the space of one of the scanners used in the model's training, embedding the unique characteristics of the selected scanner into the transferred image.
A key strength of our model lies in its ability to generalize effectively to unseen scanners not included in the training set. We evaluated our method using MR images from healthy controls across different scanners, traveling subjects, and patients with AD. Additionally, we tested the model's performance across multiple applications, including brain age prediction, biomarkers extraction, AD classification, and diagnosis prediction. In all cases, our method demonstrated significant improvements in reliability and predictive accuracy compared to existing state-of-the-art image-based approaches, such as STGAN and IGUANe. Furthermore, our harmonization model eliminates the need for time-intensive preprocessing steps, such as skull-stripping, which can introduce errors by either removing brain tissue or retaining non-brain structures. This feature makes our approach particularly advantageous for applications requiring analysis of the entire head, such as research on head trauma or cranial deformities. Notably, our method provides a robust tool for harmonizing images without the necessity of a new training phase, allowing seamless integration into various neuroimaging workflows.

\section{Conclusions}
\label{sec:conclusions}
Our results demonstrate that the proposed harmonization model offers superior performance over existing methods, ensuring reproducible and consistent MRI-based analyses across diverse scanning environments. Importantly, our model does not require retraining when applied to new data, making it a practical and scalable solution for large-scale neuroimaging studies. By improving the harmonization of MR images at the source level rather than at the feature level, our method enhances the accuracy and reliability of downstream analyses while reducing preprocessing complexity. The ability to harmonize images without the need for a new training phase further underscores its efficiency and adaptability across different applications. Future work will focus on further optimizing the preprocessing pipeline by eliminating the bias field correction step, which would further streamline image processing. Additionally, we aim to extend our approach to other imaging modalities and anatomical regions. To enhance the model's robustness, future studies should include experiments with a broader range of pathologies, ensuring that the harmonization method remains effective across diverse clinical and research settings.

\section*{Acknowledgments}
This work has been partially supported by the Health Big Data Project (CCR-2018-23669122), funded by the Italian Ministry of Economy and Finance and coordinated by the Italian Ministry of Health and the network Alliance Against Cancer. The authors also acknowledge the support from the Italian Ministry of Health, with the project NeuroArtP3 (NET-2018-12366666).
All authors acknowledge the following grants by the Italian Ministry of Health (RRC-2016-2361095; RRC-2017-2364915; RRC-2018-2365796; RCR-2019-23669119\_001; RCR 2020-23670067; RCR 2022-23682285) and by the Ministry of Economy and Finance (CCR-2017-23669078) for the data acquisition.
L. Caldera is funded by Health Big Data project sponsored by the Italian Ministry of Health (CCR-2018-23669122). L. Cavinato is funded by the National Plan for NRRP Complementary Investments “Advanced Technologies for Human-centred Medicine” (PNC0000003). L. Caldera, L. Cavinato and F. Ieva acknowledge the MIUR Excellence Department Project 2023-2027 awarded to Dipartimento di Matematica, Politecnico di Milano.

\subsection*{RIN IRCCS List} 
The RIN Neuroimaging Network is constituted by the following centers: IRCCS Istituto Auxologico Italiano (Milan); IRCCS Ospedale pediatrico Bambino Gesù (Rome); Fondazione IRCCS Istituto neurologico “Carlo Besta” (Milan); IRCCS Centro Neurolesi “Bonino Pulejo” (Messina); Centro IRCCS “Santa Maria nascente” - Don Gnocchi (Milan); IRCCS Istituto Centro San Giovanni di Dio Fatebenefratelli (Brescia); IRCCS Ospedale pediatrico “Giannina Gaslini” (Genoa); IRCCS Istituto Clinico Humanitas (Milan); Istituto di Ricerche Farmacologiche Mario Negri IRCCS (Milan); Istituti Clinici Scientifici Maugeri, IRCCS (Pavia); IRCCS Eugenio Medea (Bosisio Parini); Fondazione IRCCS Istituto Neurologico “Casimiro Mondino” (Pavia); IRCCS NEUROMED – Istituto Neurologico Mediterraneo (Pozzilli); IRCCS Associazione Oasi Maria SS Onlus – Troina (Enna); Fondazione IRCCS Ca’ Granda Ospedale Maggiore Policlinico (Milan); IRCCS Fondazione Ospedale San Camillo (Venice); IRCCS Ospedale San Raffaele (Milan); IRCCS Fondazione Santa Lucia (Rome); IRCCS Istituto di Scienze Neurologiche (Bologna); IRCCS SDN Istituto di ricerca diagnostica e nucleare (Naples); IRCCS Fondazione Stella Maris (Pisa); IRCCS San Martino (Genova); IRCCS Gemelli (Roma).

\subsection*{Code and Data Availability}
The implementation of the proposed approach is accessible at this \href{https://github.com/luca2245/DISARMpp_Harmonization.git}{link}. ADNI3, PPMI, SALD, IXI, SRPBS are publicy available datasets. The private datasets that were used for testing in this study are available from the Italian Neuroimaging Network (RIN) and Ospedale Policlinico San Martino (NeurArtP3) but restrictions apply to the availability of these data, which were used under license for the current study, and so they are not publicly available. Data are however available from the authors upon reasonable request and with permission of both  Italian Neuroimaging Network (RIN) and Ospedale Policlinico San Martino.

\newpage

\section*{\textbf{\Large{Appendix}}}
\appendix

\section*{Appendix A: Detailed Description of the MRI Datasets}
\label{appx:app_A}
This section presents a comprehensive overview of the MR T1-weighted images used in the main manuscript. Tables \ref{tab:datasets} and \ref{tab:AD_datasets} provide detailed descriptions of the datasets for healthy controls and Alzheimer's disease (AD) patients, respectively. These tables include essential information such as scanner types, manufacturers, the number of images, field strengths, and participant age ranges.
We obtained T1-weighted MR images of healthy controls from five distinct datasets. Specifically, the collection comprises 313 images from the Alzheimer's Disease Neuroimaging Initiative (ADNI3), acquired using five different scanners; 60 images from the Parkinson's Progression Markers Initiative (PPMI), obtained with three scanners; 581 images from the IXI Brain Development Dataset (IXI), acquired with three scanners; 494 images from the Southwest University Adult Lifespan Dataset (SALD), captured using a single scanner; and 117 images from a private dataset provided by the Italian Neuroimaging Network (RIN), collected with six different scanners.
For AD patients, the dataset includes 41 MR scans from the private NeuroArtP3 dataset and 61 MR scans from the public ADNI3 dataset. Additionally, we include data from seven subjects in the SRPBS traveling subjects dataset \cite{tanaka2021multi}. Table \ref{tab:travel_sub_dataset} provides a detailed breakdown of this dataset, including the number of sites, scanner models, field strengths, and demographic information (age and sex) of the subjects. The "Number of Sites" column represents the locations where each subject underwent MRI scans, corresponding to the total number of images considered per subject. In the "Scanner Models" column, the numbers in parentheses indicate the count of images obtained using each scanner model across different sites. 

\begin{table}[H]
\scriptsize
\centering
\begin{tblr}{@{} >{\raggedright\arraybackslash} m{1.4cm} >{\raggedright\arraybackslash}m{2.9cm} >{\raggedright\arraybackslash}m{2.0cm} >{\centering\arraybackslash}m{1.3cm} >{\centering\arraybackslash}m{1.1cm} >{\centering\arraybackslash}m{1.4cm} >{\centering\arraybackslash}m{1.4cm} @{}}
\hline[0.5pt]
\textbf{Dataset} & \textbf{Scanner Model} & \textbf{Manufacturer} & \textbf{Img. \#} & \textbf{Total} & \textbf{Field Strength} & \textbf{Age Range} \\  
\hline \hline
\SetCell[r=5]{c,1.5cm} \textbf{ADNI3} \cite{jack2008alzheimer} & Achieva dStream & Philips & 26  & \SetCell[r=5]{c, 1.3cm} 313 & \SetCell[r=5]{c,1.5cm} 3.0T & \SetCell[r=5]{c,1.5cm} 50-95 \\
&  Prisma Fit & SIEMENS & 167 & & &  \\
&  Prisma & SIEMENS & 69 & & & \\
&  Skyra & SIEMENS & 38 & &  & \\
&  Achieva & Philips & 13 & & & \\
\hline
\SetCell[r=3]{c,1.5cm} \textbf{PPMI} \cite{marek2011parkinson} & Achieva dStream & Philips & 5  & \SetCell[r=3]{c,1.3cm} 60 & \SetCell[r=3]{c,1.5cm} 3.0T & \SetCell[r=3]{c,1.5cm} 60-80 \\
&  Achieva & Philips & 14 & & &  \\
&  Triotim & SIEMENS & 41 & &  & \\
\hline
\SetCell[r=3]{c,1.5cm} \textbf{IXI} \cite{ixidata} & Gyroscan Intera & Philips & 322  & \SetCell[r=3]{c,1.3cm} 581 &  1.5T & \SetCell[r=3]{c,1.5cm} 20-86 \\
&  Intera & Philips & 185 & & 3.0T &  \\
&  Unknown & GE & 74 & & 1.5T  & \\
\hline
\SetCell[r=1]{c,1.5cm} \textbf{SALD} \cite{wei2017structural} & Trio & SIEMENS & 494  & \SetCell[r=1]{c,1.3cm} 494 &  3.0T & \SetCell[r=1]{c,1.5cm} 19-80 \\
\hline
\SetCell[r=6]{c,1.5cm} \textbf{RIN} \cite{nigri2022quantitative} & Achieva dStream & Philips & 17 & \SetCell[r=6]{c,1.3cm} 115 & \SetCell[r=6]{c,1.5cm} 3.0T & \SetCell[r=6]{c,1.5cm} 49-80 \\
&  DISCOVERY MR750 & GE & 20 & & &  \\
&  Ingenia CX & Philips & 23 & & & \\
&  Prisma & SIEMENS & 40 & &  & \\
&  Skyra & SIEMENS & 12 & & & \\
&  Skyra Fit & SIEMENS & 3 & & & \\
\hline[0.5pt]
\end{tblr}
\caption{Healthy controls datasets description, including scanner types, manufacturers, number of images, field strengths, and participant age range.}
\label{tab:datasets}
\end{table}

\begin{table}[H]
\scriptsize
\centering
\begin{tblr}{@{} >{\raggedright\arraybackslash} m{1.7cm} >{\raggedright\arraybackslash}m{2.9cm} >{\raggedright\arraybackslash}m{2.0cm} >{\centering\arraybackslash}m{1.3cm} >{\centering\arraybackslash}m{1.1cm} >{\centering\arraybackslash}m{1.4cm} >{\centering\arraybackslash}m{1.4cm} @{}}
\hline[0.5pt]
\textbf{Dataset} & \textbf{Scanner Model} & \textbf{Manufacturer} & \textbf{Img. \#} & \textbf{Total} & \textbf{Field Strength} & \textbf{Age Range} \\  
\hline \hline
\SetCell[r=6]{c,1.5cm} \textbf{NeuroArtP3} \cite{malaguti2024artificial}  & Achieva & Philips & 17  & \SetCell[r=6]{c, 1.3cm} 41 & \SetCell[r=6]{c,1.5cm} 1.5T & \SetCell[r=6]{c,1.5cm} 63-81 \\
&  Genesis Signa & GE & 6 & & &  \\
&  Signa HDxt & GE & 14 & & & \\
&  Ingenia & Philips & 2 & &  & \\
&  Intera & SIEMENS & 1 & & & \\
&  Magnetom Espree & SIEMENS & 1 & & & \\
\hline
\SetCell[r=10]{c,1.5cm} \textbf{ADNI3} \cite{jack2008alzheimer} & Prisma & SIEMENS & 9  & \SetCell[r=10]{c,1.3cm} 61 &  \SetCell[r=10]{c,1.5cm} 3.0T & \SetCell[r=10]{c,1.5cm} 55-89 \\
& Ingenia & Philips & 5 & & / &  \\
& Prisma Fit & SIEMENS & 20 & & /  & \\
& Biograph mMR & SIEMENS & 2 & & /  & \\
& Skyra & SIEMENS & 11 & & /  & \\
& Triotim & SIEMENS & 3 & & /  & \\
& Ingenia Elition X & Philips & 1 & & /  & \\
& Verio & SIEMENS & 5 & & /  & \\
& Achieva & Philips & 1 & & /  & \\
& Achieva dStream & Philips & 4 & & /  & \\
\hline[0.5pt]
\end{tblr}
\caption{Alzheimer's Disease (AD) patients datasets description, including scanner types, manufacturers, number of images, field strengths, and participant age range.}
\label{tab:AD_datasets}
\end{table}

\begin{table}[H]
\scriptsize
\centering
\begin{tblr}{@{} >{\raggedright\arraybackslash} m{1.5cm}  >{\centering\arraybackslash} m{1.0cm}  >{\centering\arraybackslash} m{1.2cm}  >{\raggedright\arraybackslash} m{4.0cm} >{\centering\arraybackslash} m{1.2cm} >{\centering\arraybackslash} m{1.0cm}  >{\centering\arraybackslash} m{1.0cm} @{}}
\hline[0.5pt]
 \textbf{Dataset} & \textbf{Subject} & \textbf{No. Sites} & \textbf{Scanner Models} & \textbf{Field Strength}  &  \textbf{Age} & \textbf{Sex} \\
\hline \hline
\SetCell[r=7]{c,1.5cm} \textbf{SRPBS} \cite{tanaka2021multi} & 1 & 10 & Triotim, Verio (2), Spectra, Signa HDxt, Achieva (3), DISCOVERY MR750w, Skyra  & 3T & 25 & M \\
 & 2 & 10 & Triotim (2), Verio (2), Spectra, Signa HDxt, Skyra, DISCOVERY MR750w, Achieva (2)  & 3T & 27 & M \\
 & 3 & 9 & Triotim, Verio (2), Spectra, Signa HDxt, Achieva (3), Skyra & 3T & 26 & M \\
 & 6 & 10 & Triotim (2), Verio (2), Spectra, Signa HDxt, Achieva (3), Skyra & 3T & 24 & M \\
 & 7 & 8 & Triotim, Verio, Spectra, Signa HDxt, Achieva (3), Skyra & 3T & 25 & M \\
 & 8 & 11 & Triotim (2), Verio (2), Spectra, Signa HDxt, Achieva (3), DISCOVERY MR750w, Skyra  & 3T & 28 & M \\
 & 9 & 10 & Triotim, Verio (2), Spectra, Signa HDxt, Achieva (3), DISCOVERY MR750w, Skyra  & 3T & 30 & M \\  
\hline[0.5pt]
\end{tblr}
\caption{Description of the subjects from the SRPBS \cite{tanaka2021multi} traveling subject dataset, including the number of sites, scanner models, field strength, age, and sex. The number of sites column ("No. Sites") indicates the locations where each subject underwent MRI scans, corresponding to the total number of images considered for each subject. In the "Scanner Models" column, numbers in parentheses represent the count of images acquired with the specified scanner model across different sites.}
\label{tab:travel_sub_dataset}
\end{table}

\section*{Appendix B: Additional Details on Training Procedure}
\label{subappx:add_on_arch}
\noindent
Table \ref{table_weights} provides details on the loss weights used during model training, the dimensions of the spaces $\mathcal{B}$ and $\mathcal{S}$, and the number of training iterations.

\begin{table}[H]
\small
\centering
\begin{tblr}{@{} >{\centering\arraybackslash} m{0.5cm} >{\centering\arraybackslash}m{0.4cm} >{\raggedright\arraybackslash}m{0.5cm} >{\centering\arraybackslash}m{0.5cm} >{\centering\arraybackslash}m{0.5cm} >{\raggedright\arraybackslash}m{0.5cm} >{\centering\arraybackslash}m{0.5cm} >{\centering\arraybackslash}m{0.5cm} >
{\centering\arraybackslash}m{1.3cm} >{\centering\arraybackslash}m{1.3cm} >{\centering\arraybackslash}m{3.2cm} @{}}
\hline[0.5pt]
\textbf{$\lambda_{\text{cc}}$} & \textbf{$\lambda_{\text{rec}}$} & \textbf{$\lambda_{\text{lat}}$} & \textbf{$\lambda_{\text{KL}}$} & \textbf{$\lambda_{\text{sf}}$} &  \textbf{$\lambda_{\text{adv}}^{\text{b}}$} & \textbf{$\lambda_{\text{cls}}^{\text{s}}$} & 
\textbf{$\lambda_{\text{adv}}^{\text{s}}$} &
\textbf{dim($\mathcal{B}$)} & \textbf{dim($\mathcal{S}$)} & \textbf{Number of Iterations} \\
\hline \hline
$10$ & $10$ & $8$ & $0.01$ & $7$ & $1$ & $3$ & $10$ & $(91, 109, 91)$ & $16$ & $69$k \\
\hline[0.5pt]
\end{tblr}
\caption{Implementation details.}
\label{table_weights}
\end{table}

\section*{Appendix C: Additional Details on the Experiments}
\label{subappx:metrics_and_stat_tets}
\noindent
In this section, we provide additional details regarding the experiments discussed in Section 5 of the main manuscript.

\subsection*{Detailed Description of the Evaluation Metrics and Statistical Tests}
\label{subappx:metrics}
\noindent
We outline the metrics and statistical tests used to evaluate the performance of harmonization methods and describe how they are applied.

\subsubsection*{Anatomical Structure Metrics}
\label{subsubappx:struct_metrics}
\noindent
 For the assessment of anatomical structure preservation and the quality of generated images, we employ four metrics: (1) the structural component of the SSIM index \cite{wang2004image}, referred to as Struct-SSIM, the complete SSIM index \cite{wang2004image}, the Learned Perceptual Image Patch Similarity (LPIPS) metric \cite{zhang2018unreasonable}, and the Fréchet Inception Distance (FID) \cite{heusel2017gans}.

\subsubsection*{Complete SSIM and Structural SSIM}
\noindent
The complete SSIM index \cite{wang2004image} evaluates similarity considering luminance, contrast, and structure. A higher SSIM score indicates greater overall similarity. In our context, it is used to measure the similarity between harmonized images in the traveling subject dataset. The structural component of the SSIM index \cite{wang2004image}, referred to as Struct-SSIM, focuses on assessing the similarity of structural patterns between the reference and generated images by comparing local spatial structures independently of luminance and contrast, evaluating the preservation of fine details and spatial coherence. A higher Struct-SSIM score indicates greater structural similarity. In our context, it is used to measure the structural similarity between the original images and their harmonized versions. Since these metrics are defined for 2D images, to assess the similarity between two 3D MR scans, we calculate the metric for each slice in all three directions (sagittal, coronal, axial) and then average the results to obtain the overall similarity of the two 3D MR scans. The mathematical definitions of SSIM and Struct-SSIM are given as follows:
\begin{align}
   & \text{SSIM}(x, y) = \frac{(2\mu_x\mu_y + C_1)(2\sigma_{xy} + C_2)}{(\mu_x^2 + \mu_y^2 + C_1)(\sigma_x^2 + \sigma_y^2 + C_2)}; \\ 
   & \text{Struct-SSIM}(x, y) = \frac{\sigma_{xy} + C_3}{\sigma_x \sigma_y + C_3}
\end{align}
where \(\mu_x\) and \(\mu_y\) are the means, \(\sigma_x^2\) and \(\sigma_y^2\) are the variances, and \(\sigma_{xy}\) is the covariance between images \(x\) and \(y\). The terms \(C_1\), \(C_2\), and \(C_3\) are small constants for numerical stability.

\subsubsection*{Learned Perceptual Image Patch Similarity (LPIPS)}
\noindent
The Learned Perceptual Image Patch Similarity (LPIPS) metric \cite{zhang2018unreasonable}, which measures perceptual similarity between two images by comparing feature activations from a deep neural network, making it sensitive to semantic and textural differences and aligned with human visual perception. A lower LPIPS score indicates higher similarity. As the first index, we use the LPIPS to evaluate the perceptual similarity between pairs of original and harmonized images. Specifically, since LPIPS is defined for RGB 2D images, the comparison is performed by extracting 2D slices from the MRI volumes and converting them to RGB by replicating the single channel across the three color channels. Finally, given the computational intensity of LPIPS, we limit comparisons to a subset $B$ of central slices for each of the $T$ images considered. This process is applied across axial, sagittal, and coronal orientations, and the final LPIPS value is the average across all $3 \times B \times T$ slice comparisons. The LPIPS formula is given by:
\begin{align}
\text{LPIPS}(x, y) = \sum_l \frac{1}{H_l W_l} \sum_{h,w} \| w_l \big( f^l(x)_{hw} - f^l(y)_{hw} \big) \|^2  
\end{align}
where \( f^l(x) \) and \( f^l(y) \) are the deep features at layer \( l \) for images \( x \) and \( y \), \( w_l \) are the learned weights, and \( H_l \) and \( W_l \) denote the height and width of the feature map at layer \( l \).

\subsubsection*{Fréchet Inception Distance (FID)}
\noindent
The Fréchet Inception Distance (FID) \cite{heusel2017gans}, which quantifies the distributional similarity between real and generated images by comparing their feature embeddings extracted from a pre-trained Inception network, effectively assessing the overall quality and realism of the generated images, with lower values indicating closer alignment to the real data distribution. Similar to the LPIPS, the FID is defined for RGB 2D images, so we apply the same conversion process. However, since FID compares sets of real and generated images and outputs a single scalar value, we compute it by comparing the set of original images with the set of harmonized images. Specifically, we extract $B$ central slices from each of the $T$ images considered for both the original and harmonized sets. This results in two sets, each containing $B \times T$ slices. The FID is computed by comparing these sets separately for the axial, sagittal, and coronal orientations. Finally, the overall FID score is obtained by averaging the FID values from all three orientations. The FID formula is given by:
\begin{align}
\text{FID} = \| \mu_r - \mu_g \|^2 + \text{Tr}\bigg( \Sigma_r + \Sigma_g - 2\sqrt{\Sigma_r\Sigma_g} \bigg)
\end{align}
where \(\mu_r\) and \(\mu_g\) represent the means, and \(\Sigma_r\) and \(\Sigma_g\) are the covariances of the real and generated image features.

\subsubsection*{Harmonization Metrics}
\label{subsubappx:harm_metrics}
\noindent
To evaluate the transfer of scanner characteristics, we assessed the similarity of voxel intensity distributions before and after harmonization using three metrics: Jensen-Shannon Divergence (JSD) \cite{lin1991divergence}, Hellinger Distance (HD) \cite{kailath2003divergence, rao1995review}, and Wasserstein Distance (WD) \cite{villani2008optimal}. Practically, given the MRI scans having dimension \( (1, H, W, D) \) — where \( H \) represents height, \( W \) represents width, and \( D \) represents depth — we computed these metrics based on the set of $1 \times H \times W \times D$ voxel intensity distributions derived from MRI scans, considering pre- and post-harmonization data. Specifically, we compute the empirical distribution by estimating the underlying probability distribution of the voxel intensities for each unfolded image, and we average the distributions of images belonging to the same scanner. In this way, we obtain  \( G \) distributions — with \( G \) being the number of the test scanners — from pre-harmonization images and \( G \) distributions from post-harmonization images. In each of the two sets of $G$ distributions, we computed the values for the three metrics across all possible pairs to quantify the similarity between them. Therefore, when reporting these metrics, we present the mean and standard deviation of all pairwise comparison values, providing a general assessment for all scanners.  We perform this step for both pre-harmonization and post-harmonization sets, enabling the comparison of the similarity between the distributions before and after harmonization. To statistically evaluate the effectiveness of harmonization, we perform a paired t-test comparing the values of the similarity metrics before and after harmonization. This test assesses whether the mean difference is significantly different from zero. If the differences do not follow a normal distribution, we employ a bootstrap paired t-test, resampling the data to estimate the distribution of differences. We report the 95\% confidence interval (CI) for these differences; if the CI does not include zero, it indicates a significant effect of harmonization on voxel intensity similarity.

Besides the three aforementioned indexes, in certain analyses, we also employed the K-sample Anderson-Darling test (AD-test) \cite{Anderson-Darling}, a non-parametric test that evaluates whether multiple samples originate from the same distribution. This test was applied to the set of \( G \) distributions, separately for pre- and post-harmonization images. If we accept the null hypothesis, it suggests no significant difference between distributions, indicating successful harmonization. Conversely, rejecting the null hypothesis implies that at least one distribution differs, suggesting incomplete harmonization.

\subsubsection*{Jensen-Shannon Divergence (JSD)}
\noindent
The Jensen–Shannon Divergence (JSD)~\cite{lin1991divergence} is a symmetric and bounded metric that quantifies the difference between two probability distributions. Unlike the Kullback–Leibler (KL) divergence, which is asymmetric, the JSD improves upon it by symmetrizing the measure. This is achieved by computing the KL divergence of each distribution relative to their average distribution \( M = \frac{1}{2}(P + Q) \), and then taking the mean of these two values. Formally, the JSD is defined as the average of the KL divergences from \( P \) and \( Q \) to \( M \) as:
\begin{align}
\text{JSD}(P, Q) = \frac{1}{2} \text{KL}(P \| M) + \frac{1}{2} \text{KL}(Q \| M)
\end{align}
where \( \text{KL}(P \| M) \) is the Kullback-Leibler divergence between \( P \) and \( M \). The JSD is bounded between 0 and \( \log(2) \), where:
\begin{itemize}
    \item \( \text{JSD}(P, Q) = 0 \) indicates that the two distributions are identical.
    \item \( \text{JSD}(P, Q) = \log(2) \) indicates that the distributions are maximally different.
\end{itemize}
Because of its symmetry and bounded nature, the JSD is often preferred for comparing distributions when assessing their similarity or dissimilarity, especially when dealing with probability distributions that may have different supports or shapes. It has strong interpretability and guarantees a finite value even when distributions exhibit significant differences.

\subsubsection*{Hellinger Distance (HD)}
\noindent
The Hellinger Distance (HD) is a metric derived from the Bhattacharyya coefficient, quantifying the similarity between two probability distributions by focusing on the overlap of their supports. It is particularly useful in geometric comparisons, measuring the distance between two distributions in terms of their density overlap. The Hellinger distance is defined as:
\begin{align}
\text{HD}(P, Q) = \frac{1}{\sqrt{2}} \sqrt{\sum_{x} \left( \sqrt{P(x)} - \sqrt{Q(x)} \right)^2}
\end{align}
where \( \sqrt{P} \) and \( \sqrt{Q} \) represent the square roots of the probability densities \( P \) and \( Q \). The Hellinger distance ranges from 0 to 1:
\begin{itemize}
    \item \( \text{HD}(P, Q) = 0 \) indicates that the two distributions are identical.
    \item \( \text{HD}(P, Q) = 1 \) indicates that the distributions are completely disjoint.
\end{itemize}
One of the strengths of the Hellinger distance is its sensitivity to differences in the tails of distributions. This sensitivity can reveal subtle distinctions that might be overlooked by other metrics. As such, the Hellinger distance is often preferred in cases where small differences in tail behavior are of importance.

\subsubsection*{Wasserstein Distance (WD)}
\noindent
The Wasserstein Distance (WD) measures the minimal cost of transforming one distribution into another. This cost is computed based on how much mass needs to be moved and how far it must be transported to convert one distribution into another. The Wasserstein distance considers both the magnitude of differences between distributions and the spatial arrangement of these differences. Mathematically, the Wasserstein distance is defined as the infimum of the total transportation cost over all possible couplings between the two distributions \( P \) and \( Q \):
\begin{align}
\text{WD}(P, Q) = \int_{\mathbb{R}} \left| F_P(x) - F_Q(x) \right| dx
\end{align}
where \( F_P(x) \) and \( F_Q(x) \) are the cumulative distribution functions of \( P \) and \( Q \), respectively. The Wasserstein distance is sensitive to the geometry of distributions, making it particularly useful when the spatial arrangement of data is crucial. The Wasserstein distance is unbounded and can take arbitrarily large values, especially when the supports of the two distributions are far apart. 

Using all three metrics — Jensen-Shannon Divergence (JSD), Hellinger Distance (HD), and Wasserstein Distance (WD) —  enables a comprehensive and nuanced assessment of probability distribution similarity. Each metric highlights different aspects of distributional differences:

\begin{itemize}
    \item \textbf{JSD}: Offers a symmetric, bounded measure of divergence, effective for general similarity comparisons but may overlook subtle differences in distribution tails.
    \item \textbf{HD}: Provides sensitivity to small discrepancies, especially in tails, making it valuable for geometric structure comparisons.
    \item \textbf{WD}: Captures minimal transformation cost between distributions, considering both magnitude and spatial arrangement, ensuring deeper interpretability.
\end{itemize}
By leveraging all three metrics, a more holistic perspective on the similarities and dissimilarities of probability distributions can be obtained. This multi-metric approach ensures a robust comparison, especially in complex analyses where a single measure may not fully capture all relevant information.

\subsection*{Additional details on the Downstream Analysis}
\label{subappx:down}
\noindent
This section offers further details about the downstream analysis procedures.

\subsubsection*{Inter-Scanner Variability in MRI-Derived Brain Volumes}
\label{subsubappx:inter-scan}
\noindent
In the analysis, we employed Linear Mixed-effects Models (LMM) to predict these volumes based on age, treating scanner groups as random effects. 
The LMM is defined as follows:
\[
y_{ij} = \beta_0 + \beta_1 x_{ij} + u_j + \epsilon_{ij}
\]
where \( y_{ij} \) and \( x_{ij} \) represent the age and volume, respectively, for individual \( i \) scanned using scanner \( j \). Here, \( \beta_0 \) is the fixed intercept, and \( \beta_1 \) is the fixed-effect coefficient for volume. The term \( u_j \) represents the random effect for scanner group \( j \), assumed to be normally distributed with variance \( \sigma_u^2 \). \( \epsilon_{ij} \) is the residual error term, assumed to be normally distributed with variance \( \sigma_\epsilon^2 \). The random effect \( u_j \) captures between-group variance, whereas the residual error \( \epsilon_{ij} \) captures within-group variance.
The ICC and $R^2_m$ are computed as:
\[
\text{ICC} = \frac{\sigma_u^2}{\sigma_u^2 + \sigma_\epsilon^2}; \quad \quad R^2_m = \frac{\sigma_{\beta}^2}{\sigma_{\beta}^2 + \sigma_{\epsilon}^2 + \sigma_u^2}
\]
where $\sigma_{\beta}^2$ is the variance explained by the fixed effects.

\section*{Appendix D: Additional Details on the Results}
In this section, we present supplementary information and visualizations related to the results discussed in Section 6 of the main manuscript.

\subsection*{Visual Assessment of Harmonization}
\label{subappx:visual_assess}
\begin{figure}[H]
    \includegraphics[width=\linewidth]{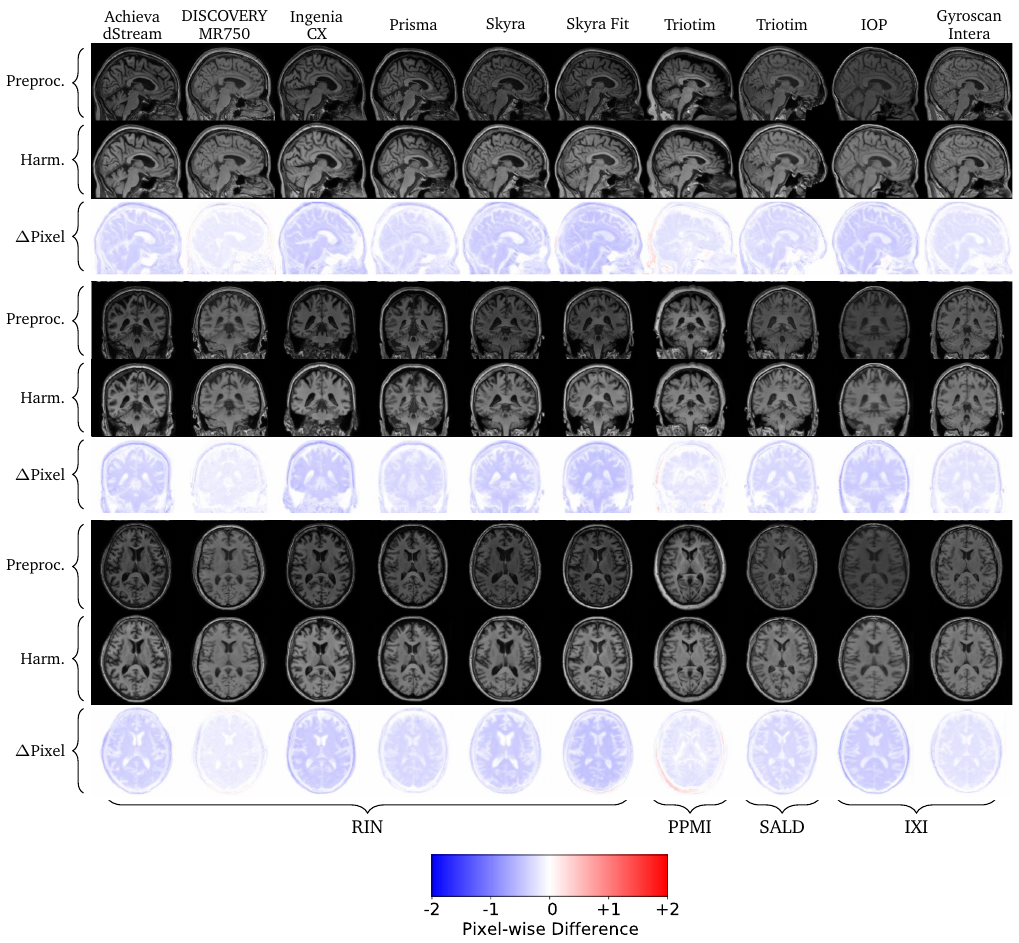} 
        \caption{DISARM++ visual assessment of the harmonization to Gyroscan Intera reference scanner space. The figure displays slices from the axial, coronal, and sagittal dimensions for 10 original images—one per test scanner—alongside their corresponding harmonized slices. Heatmaps illustrate the pixel-wise differences between the harmonized images and their original counterparts.}
    \label{fig:disarm_harm_imgs_with_heatmaps_gi}
\end{figure}

In this section, we provide further details related to Section 6.2.1 of the main manuscript. Figure \ref{fig:disarm_harm_imgs_with_heatmaps_gi} displays 10 original images (one from each test scanner), showing slices from axial, coronal, and sagittal dimensions. These slices are presented alongside their harmonized counterparts with DISARM++ to the Gyroscan Intera reference scanner space.

\subsection*{Assessment of Anatomical Structure Preservation and Scanner Characteristics Transfer}
\begin{figure}[H]
        \centering
        \includegraphics[width=\linewidth]{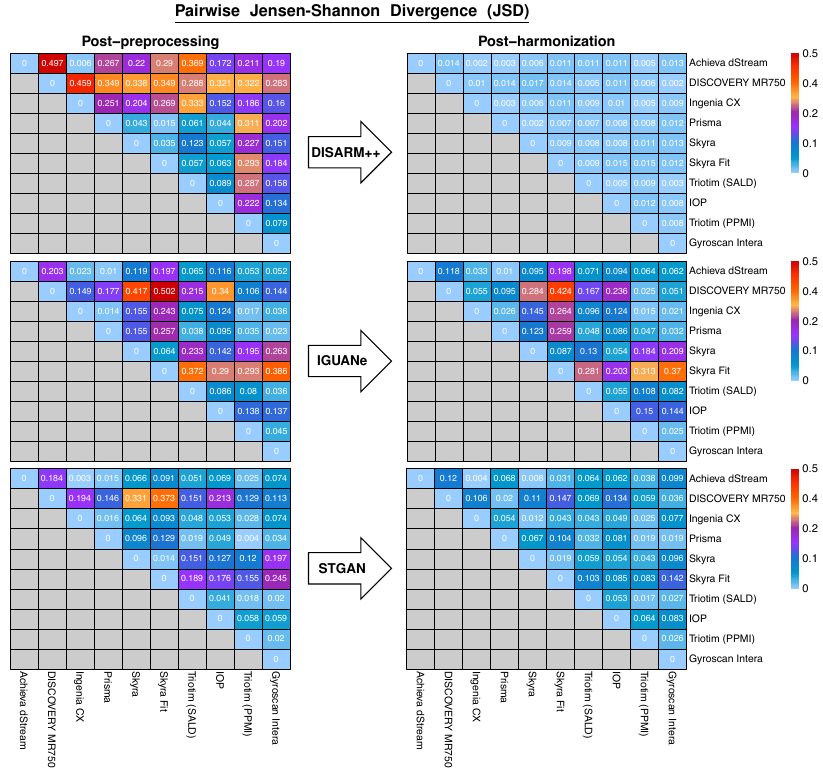}
        \caption{Heatmaps illustrating the Jensen-Shannon Divergence (JSD) between the mean distributions of each test scanner pair, both after preprocessing and post-harmonization using DISARM++, IGUANe, and STGAN, where lower values indicate greater similarity.}
        \label{fig:jsd_comparison}
\end{figure}

\begin{figure}[H]
        \centering
        \includegraphics[width=\linewidth]{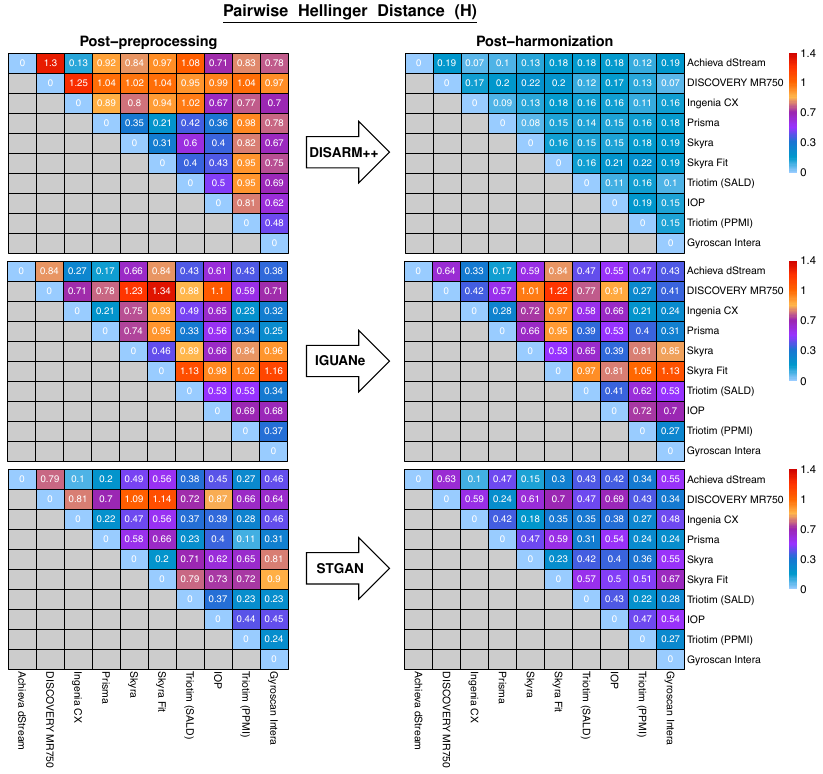}
        \caption{Heatmaps illustrating the Hellinger Distance (H) between the mean distributions of each test scanner pair, both after preprocessing and post-harmonization using DISARM++, IGUANe, and STGAN, where lower values indicate greater similarity.}
        \label{fig:hell_comparison}
\end{figure}

\begin{figure}[H]
        \centering
        \includegraphics[width=\linewidth]{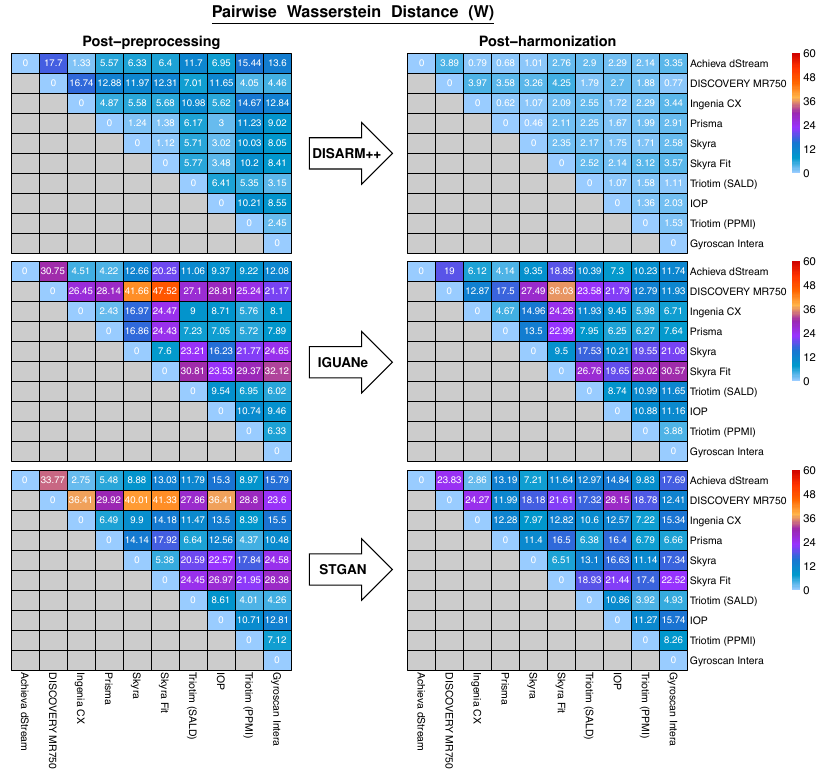}
        \caption{Heatmaps illustrating the Wasserstein Distance (W) between the mean distributions of each test scanner pair, both after preprocessing and post-harmonization using DISARM++, IGUANe, and STGAN, where lower values indicate greater similarity.}
        \label{fig:wass_comparison}
\end{figure}

In this section, we provide additional details related to Section 6.2.2 of the main manuscript. Figures \ref{fig:jsd_comparison}, \ref{fig:hell_comparison}, and \ref{fig:wass_comparison} present heatmaps illustrating all pairwise values for DISARM++, IGUANE, and STGAN before and after harmonization across the three metrics. Each cell in the heatmaps represents the similarity between the mean voxel intensity distributions of a pair of scanners.

\newpage

\bibliographystyle{apalike}
\bibliography{bibliography}

\end{document}